\documentclass[10pt,journal,compsoc]{IEEEtran}
\usepackage{multirow}

\usepackage[breaklinks=true,colorlinks,bookmarks=false]{hyperref}
\newlength\savewidth

\ifCLASSOPTIONcompsoc
  \usepackage[nocompress]{cite}
\else
  \usepackage{cite}
\fi
\usepackage{bbding}
\usepackage[pdftex]{graphicx}
\ifCLASSINFOpdf
\else
\fi
\usepackage{amsmath}
\usepackage{algorithm}
\usepackage{algorithmic}
\usepackage{array}

\usepackage{url}
\begin{document}
%
\title{From Handcrafted to Deep Features for Pedestrian Detection: A Survey}
%
%
%
%

\author{Jiale Cao,
        Yanwei Pang,~\IEEEmembership{Senior Member,~IEEE,}
        Jin Xie,
        Fahad Shahbaz Khan,~\IEEEmembership{Senior Member,~IEEE,}
        and~Ling~Shao,~\IEEEmembership{Fellow,~IEEE}
\IEEEcompsocitemizethanks{\IEEEcompsocthanksitem This work was supported in part by National Key Research and Development Program of China
(No. 2018AAA0102800) and in part by National Natural Science Foundation of China (Nos. 61906131, 61632018) and by VR starting grant (2016-05543).
\IEEEcompsocthanksitem J. Cao, Y. Pang, and J. Xie are with the School of Electrical and Information Engineering, Tianjin University, Tianjin 300072, China (E-mail: \{connor,~pyw,~jinxie\}@tju.edu.cn). (Corresponding author: Yanwei Pang)
\IEEEcompsocthanksitem F. Khan is with Mohamed bin Zayed University of Artificial Intelligence, UAE and Link{\"o}ping University, Sweden. (fahad.khan@mbzuai.ac.ae).
\IEEEcompsocthanksitem L. Shao is with the Inception Institute of Artificial Intelligence, UAE. (ling.shao@ieee.org).
}
}
%
%

\markboth{IEEE Transactions on Pattern Analysis and Machine Intelligence}%
{Cao \MakeLowercase{\textit{et al.}}: From Handcrafted to Deep Features for Pedestrian Detection: A Survey}
\IEEEtitleabstractindextext{%
\begin{abstract}
Pedestrian detection is an important but challenging problem in computer vision, especially in human-centric tasks. Over the past decade, significant improvement has been witnessed with the help of handcrafted features and deep features. Here we present a comprehensive survey on recent advances in pedestrian detection. First, we provide a detailed review of single-spectral pedestrian detection that includes handcrafted features based methods and deep features based approaches. For handcrafted features based methods, we present an extensive review of approaches and find that handcrafted features with large freedom degrees in shape and space have better performance. In the case of deep features based approaches, we split them into pure CNN based methods and those employing both handcrafted and CNN based features. We give the statistical analysis and tendency of these methods, where feature enhanced, part-aware, and  post-processing methods have attracted main attention. In addition to single-spectral pedestrian detection, we also review multi-spectral pedestrian detection, which provides more robust features for illumination variance. Furthermore, we introduce some related datasets and evaluation metrics, and a deep experimental analysis. We conclude this survey by emphasizing open problems that need to be addressed and highlighting various future directions. Researchers can track an up-to-date list at \url{https://github.com/JialeCao001/PedSurvey}.
\end{abstract}

}

\maketitle

\IEEEdisplaynontitleabstractindextext

%
\IEEEpeerreviewmaketitle

\IEEEraisesectionheading{\section{Introduction}\label{secIntroduction}}

%
%
%
%
\IEEEPARstart{H}{uman}-centric computer vision tasks (\textit{e.g.,} pedestrian detection \cite{Benenson_TYPD_ECCV_2014,Dollar_Caltech_PAMI_2010,Zhang_TRHP_PAMI_2018}, person re-identification \cite{Zhao_USL_CVPR_2013,Liao_LMOR_CVPR_2014,Ye_RAE_ECCV_2018,Karanam_FMD_PAMI_2019}, person search \cite{Xiao_JDIF_CVPR_2017,Han_QGPR_CVPR_2019,Yan_LCG_CVPR_2019,Oh_PIPER_PAMI_2020}, pose estimation \cite{Ouyang_MSDL_CVPR_2014,Newell_Hourglass_ECCV_2016,Cao_MPPE_CVPR_2017,Rogez_LCR_PAMI_2020}, and face detection \cite{Yang_WiderFace_CVPR_2016,Najibi_FARPN_CVPR_2019,Li_DSFD_CVPR_2019,Ranjan_HyperFace_PAMI_2019}) have gained significant attention over the past decade. Among these tasks, pedestrian detection is one of the most fundamental tasks with a wide range of real-world-applications. In addition to its standalone value in a variety of applications (\textit{e.g.,} video surveillance and self-driving), pedestrian detection is also a prerequisite that serves as the basis for several other vision tasks (\textit{e.g.,} person re-identification and person search). For instance, both person re-identification and person search need to first accurately detect all the existing pedestrians.

\begin{figure}[t]
\centering
\includegraphics[width=0.96\linewidth]{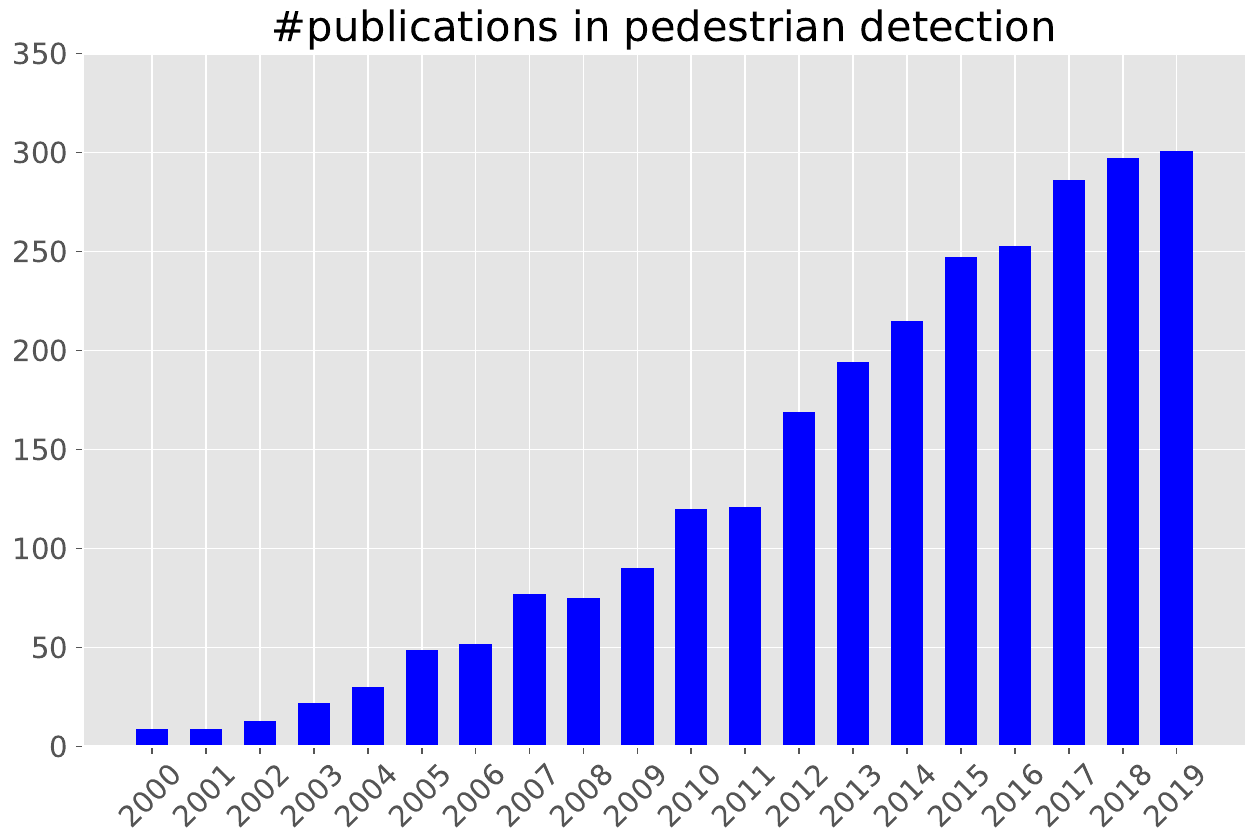}
\caption{The increasing number of publications on pedestrian detection from the year 2000 to 2019, obtained through Google scholar search with the key-words: allintitle: ``pedestrian detection''.}
\label{fig:pub}
\end{figure}

\begin{figure}[t]
\centering
\includegraphics[width=0.96\linewidth]{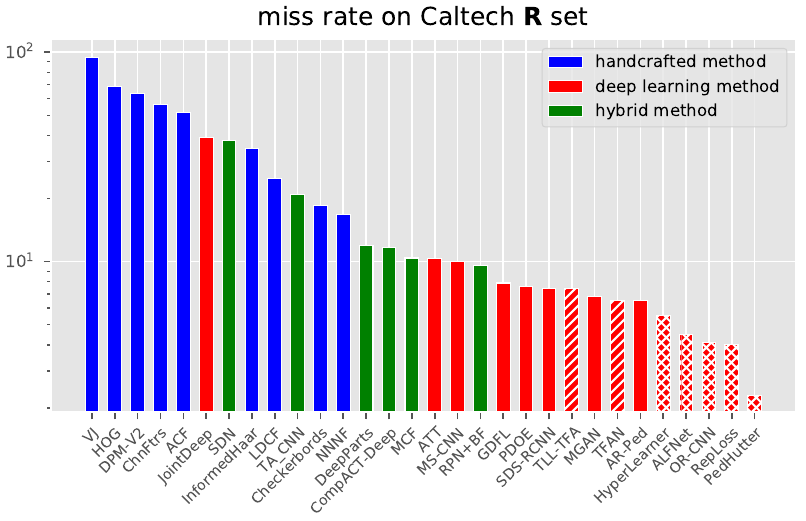}
\includegraphics[width=0.96\linewidth]{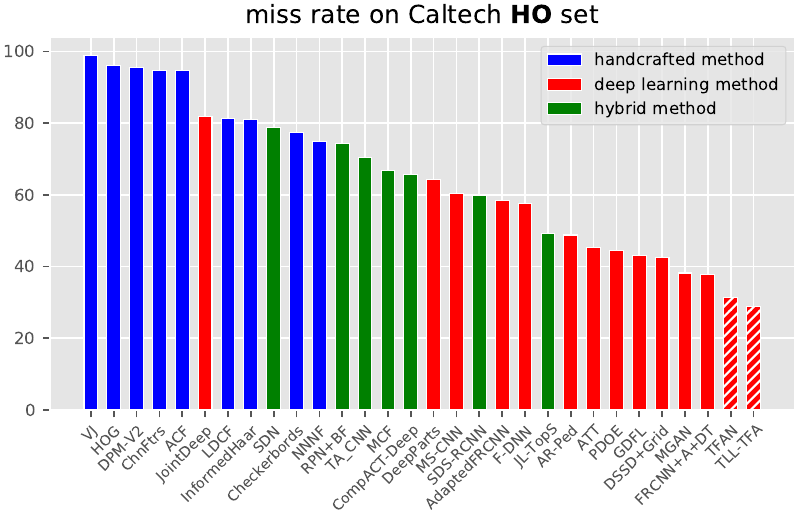}
\caption{Detection performance improvements, in terms of log-average miss rate (lower is better), on Caltech test set \cite{Dollar_Caltech_PAMI_2010} in past decade. Top: we show the performance comparison on the reasonable (\textbf{R}) set. Bottom: We show the comparison on the heavy occluded (\textbf{HO}) set. The white cross hatch in bar indicates that more accurate annotations \cite{Zhang_HowFar_CVPR_2016} are used for training and test. The white line hatch in bar indicates that motion cue is utilized in addition to appearance information.}
\label{fig:progress}
\end{figure}

The aim of pedestrian detection is to accurately localize and classify all pedestrian instances in a given image. In the past decade, pedestrian detection has received significant attention with over two thousands research publications (see Fig. \ref{fig:pub}). The increasing number of publications suggest that it is an active research problem in computer vision. In recent years, pedestrian detection performance has also obtained a consistent improvement on standard benchmarks. Fig. \ref{fig:progress} (top) shows the improvement in pedestrian detection accuracy (in terms of log-average miss rate) on the test set of Caltech \cite{Dollar_Caltech_PAMI_2010}, which is one of the most popular pedestrian detection benchmarks. The detection performance is evaluated on the reasonable \textbf{R}. The reasonable \textbf{R} set comprises pedestrians over 50 pixels in height, with less than 35\% of their body occluded. We compare the performance of 30 methods, including handcrafted, deep learning and hybrid approaches. Note that we split methods into (a) pure deep learning based approaches, comprising end-to-end training where pedestrian proposal generation and classification are learned jointly, and (b) hybrid approaches. In hybrid approaches, some methods use deep features for proposal generation and shallow classifier, such as Support Vector Machines (SVM) \cite{Cortes_SVM_ML_1995} or AdaBoost \cite{Freund_AdaBoost_ML_1997}, for proposal classification, while some other methods use handcrafted approaches for proposal generation and deep features for proposal classification. In addition, we show recent deep learning based methods (represented by white cross hatch) that utilize more accurate annotations \cite{Zhang_HowFar_CVPR_2016}. Despite the consistent progress, we argue that there is still sufficient room for improvement in order to meet real-world application requirements. For instance, Fig. \ref{fig:progress} (bottom) shows the detection performance under severe occlusions (heavy occlusion \textbf{HO} set of Caltech test dataset). The \textbf{HO} set comprises pedestrians over 50 pixels in height, with 35\% to 80\% of their body occluded. These results suggest that the detection performance under real-world challenges, such as occlusion, is still far from satisfactory.

\begin{figure*}[t]
\centering
\includegraphics[width=0.995\linewidth]{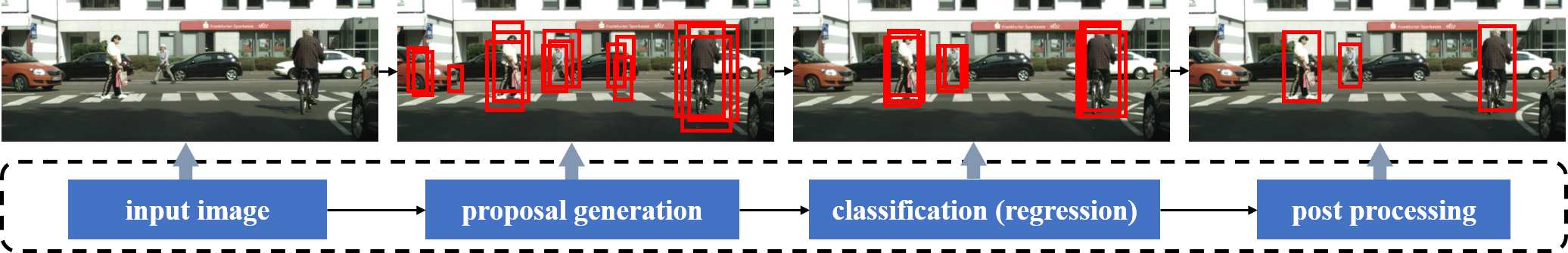}
\caption{Most pedestrian detection approaches typically comprise three consecutive steps. The first step, proposal generation, involves generating candidate proposals from an input image. The second step, proposal classification (and regression), involves assigning the proposals  to  either the  positive  class  (pedestrian)  or the negative class (background). Consequently, the post-processing step aims to suppress  duplicate bounding-boxes belonging to the same pedestrian. In proposal generation and proposal classification, feature extraction is the key. A variety of feature extraction strategies ranging from handcrafted to deep features have been used in the literature.}
\label{fig:pipeline}
\end{figure*}

\begin{figure*}[t]
\centering
\includegraphics[width=0.995\linewidth]{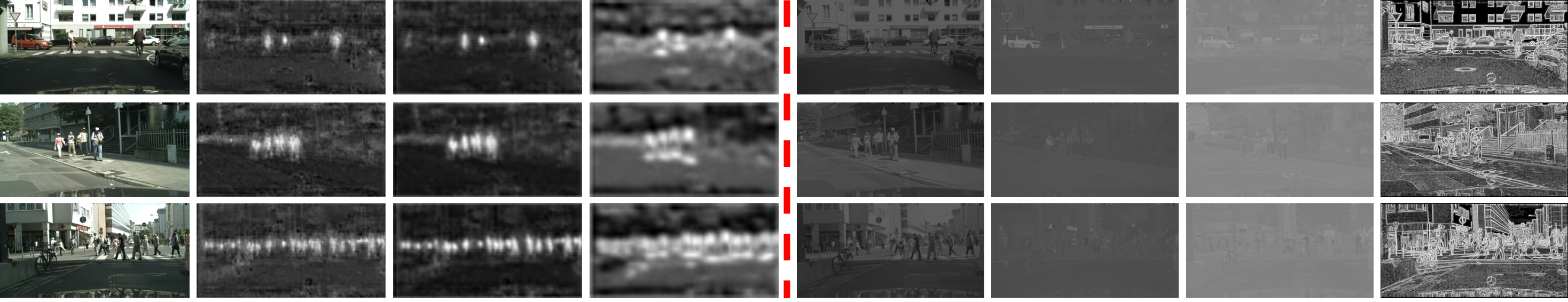}
\caption{Visualization of  the deep features and handcrafted features used in pedestrian detection. On the left (before the red dotted line): different layers (P2, P3, and P4) of the feature pyramid network \cite{Lin_FPN_CVPR_2017}. Here, we show feature channels with maximum responses. On the right (after the red dotted line): handcrafted features of three  color channels (\textit{i.e.,} LUV) followed by gradient magnitude (last column). }
\label{fig:feature}
\end{figure*}

Table \ref{table:objectdetection}  compares pedestrian detection with the related human detection and object detection. Compared with human detection, pedestrian detection primarily focuses on driving/surveillance scenes based on visible-light camera and infrared camera. In addition, pedestrian detection detects fully body, including with partial or severe occlusions, and also poses large variations in scale. Compared with generic object detection, pedestrian detection focuses on single category of pedestrian, and faces the challenges, such as frequent occurrence of partial or severe occlusions and large variations in scale. Owing to specific challenges, pedestrian detection has been studied as a standalone problem.

\begin{table}[t]
\centering
\caption{Comparison with object detection and human detection. Owing to specific challenges, pedestrian detection has been studied as a standalone problem.}
\resizebox{0.48\textwidth}{!}{
\setlength\tabcolsep{5pt}
\renewcommand\arraystretch{1.1}
\begin{tabular}{|l|c|c|c|}
\hline
Name & Pedestrian detection	 & Human detection & Object detection \\\hline\hline
Class & single & single & multiple \\
Image &  RGB/Thermal    & RGB    &  RGB \\ 
Scene & driving/surveillance & unspecific & unspecific \\
Target & full body & visible part & visible part \\
Orientation & upright & any & any \\
Occlusion frequency & large & large & medium \\
Scale variance & large & medium & medium \\
\hline
\end{tabular}}
\label{table:objectdetection}
\end{table}

In this work, we divide existing pedestrian detection works into single-spectral pedestrian detection and multi-spectral pedestrian detection. Single-spectral pedestrian detection means that only a single sensor is used for detection (\textit{e.g.,} visible-light camera and fisheye camera). Here, we mainly focus on methods based on visible-light camera. Different to single-spectral methods, multi-spectral pedestrian detection adopts multiple sensors of different types. For this class of methods, we mainly focus on those based on  visible-light and infrared cameras. Compared with single-spectral pedestrian detection, multi-spectral pedestrian detection is more robust with respect to illumination variation and has attracted considerable attention in the past few years. 

The rest of this article is organized as follows. We first introduce the detection pipeline of pedestrian detection in Section \ref{sec:pipeline}. After that, we present a detailed review and analysis of single-spectral pedestrian detection approaches  in Section \ref{sec-sspd}, including both handcrafted features based methods and deep features based approaches. Then, we introduce multi-spectral pedestrian detection in Section \ref{sec-mspd}, which is  a supplement to single-spectral pedestrian detection. An experimental analysis is provided in Section \ref{sec-data}. Finally, we discuss several existing challenges in Section \ref{sec-challenges}. 

Some surveys about pedestrian detection \cite{Geronimo_ADAS_PAMI_2009,Benenson_TYPD_ECCV_2014,Ragesh_PDAS_ACCESS_2019} have been published in past years. Compared with these previous surveys, we focus more attention on the recent deep features based methods, instead of handcrafted features based methods. Further, we present a more detailed analysis. Based on this analysis, we summarize the ongoing challenges in pedestrian detection research. We hope that this survey will not only provide a better understanding of pedestrian detection but also facilitate future research activities and various application developments in the field.

\section{Pedestrian detection pipeline}
\label{sec:pipeline}
Most pedestrian detection methods, including handcrafted \cite{Dalal_HOG_CVPR_2005,Dollar_ICF_BMVC_2009}, deep learning \cite{Zhang_Citypersons_CVPR_2017,Cai_MSCNN_ECCV_2016} and hybrid \cite{Cao_MCF_TIP_2016,Tian_DeepParts_ICCV_2015} approaches, typically comprise three consecutive steps: proposal generation, classification (and regression), and post processing. Fig. \ref{fig:pipeline} shows the overall pipeline depicting these three steps. Note that not all approaches have these three steps. For example, some approaches  \cite{Liu_ALFNet_ECCV_2018,Liu_CSP_CVPR_2019} do not have the step of proposal generation, while the recent PED \cite{Lin_PETR_arXiv_2020} does not need the NMS post-processing. Without loss of generality, we discuss these three steps in detail.

(a) \textit{Proposal generation:} This step aims to extract some candidate proposals of pedestrians from an input image. The proposals indicate a set of bounding-boxes which potentially represent the objects. Common strategies  include sliding-window methods \cite{Dalal_HOG_CVPR_2005,Viola_Haar_IJCV_2004,Dollar_ICF_BMVC_2009}, particle-window methods \cite{Gualdi_MPW_PAMI_2011,Pang_iPW_TCYB_2017}, objectness methods \cite{Zitnick_Edge_ECCV_2014,Cheng_BING_CVPR_2014,Uijlings_SS_IJCV_2013,Hosang_EDP_PAMI_2016}, and region proposal networks \cite{Ren_FasterRCNN_NIPS_2015,Cai_MSCNN_ECCV_2016,Wang_GARPN_CVPR_2019}. The sliding-window methods (SW) adopt a greedy search strategy with a fixed-sized step to scan the image from the top-left to bottom-right region. The particle-window methods adopt the coarse-to-fine cascaded search where the proposals generated at current stage follow the likelihood distribution of previous stage. The objectness methods typically employ a variety of low-level features (\textit{e.g.,} edge and color features) to extract the proposals in a bottom-up fashion. Recently, a region proposal network (RPN) was introduced for proposal generation, which shares the deep features with the following proposal classification and regression.

(b) \textit{Proposal classification:} This step assigns these  candidate proposals to the positive class (pedestrians) or the negative class (background) based on the extracted features of these proposals. The handcrafted features based methods \cite{Dalal_HOG_CVPR_2005,Benenson_TYPD_ECCV_2014} adopt a shallow classifier (\textit{e.g.,} SVM or boosting) for classification, whereas deep features based methods \cite{Ren_FasterRCNN_NIPS_2015,Liu_SSD_ECCV_2016,Dai_RFCN_NIPS_2016} generally integrate the feature extraction and classification into a unified framework by utilizing a  softmax (or sigmoid) layer. Additionally, deep features based methods add \textit{regression} in parallel with \textit{classification} to refine the location quality of the bounding-boxes.

(c) \textit{Post processing:}  As shown in Fig. \ref{fig:pipeline}, a single pedestrian may be detected by multiple bounding-boxes after proposal classification, which is the issue of duplicate detections. The technique of non-maximum suppression (NMS) selects the best bounding-box for each object and suppresses other duplicate bounding-boxes. The related methods can be divided into two categories: heuristic-based and learning-based methods. The heuristic-based methods (\textit{e.g.,} greedy NMS, Soft-NMS \cite{Bodla_SoftNMS_ICCV_2017}, SGE-NMS \cite{Yang_SGNMS_arXiv_2019}, and Adaptive NMS \cite{Liu_AdaptiveNMS_CVPR_2019}) combine the bounding-boxes according to classification scores, where the overlapped bounding-boxes with lower scores are suppressed. The learning-based methods, including Gnet \cite{Hosang_LNMS_CVPR_2017} and Relation Network \cite{Hu_RN_CVPR_2018}, learn a mapping to retain the most accurate bounding-boxes. 

\begin{table*}
\centering
\caption{Summary of 21 typical handcrafted features based methods for pedestrian detection. A shallow classifier is used to learn pedestrian detector. `CF' means channel features based method, `DPM' means deformable part model based method, and `SW' means sliding-window strategy.}
\resizebox{0.99\textwidth}{!}{
\setlength\tabcolsep{4pt}
\renewcommand\arraystretch{1.1}
\begin{tabular}{|l|l|c|c|c|c|c|c|c|c|l|}
\hline
Method & Publication & Family & Proposal & Feature & Classifier & Post-proc. & Scale-aware & Part-aware & Context & Description\\
\hline
\hline
VJ \cite{Viola_Haar_IJCV_2004} & IJCV2004 & CF & SW & RGB(haar) & boosting & NMS  & no & no &  no & a robust real-time face detector with Haar features\\
HOG \cite{Dalal_HOG_CVPR_2005} & CVPR2005 & DPM & SW & HOG & SVM & NMS & no & no  &  no & a novel histogram of gradient feature descriptor\\
HOG-LBP \cite{Yan_HOGLBP_ICCV_2010} & ICCV2009 & DPM & SW & HOG & SVM & NMS & no & yes  &  no & an occlusion likelihood map for occlusion handling\\
ChnFtrs \cite{Dollar_ICF_BMVC_2009} & BMVC2009 & CF & SW & Chntrs & boosting & NMS & no & no  & no & the simple and effective integral channel features\\
DPM \cite{Felzenszwalb_DPM_PAMI_2010} & PAMI2010 & DPM & SW & HOG & SVM & NMS & no &  yes & no & deformable part model with six parts and one root\\
HOF+CSS \cite{Walk_NF_CVPR_2010} & CVPR2010 & DPM & SW & HOG+CSS & SVM & NMS & no&  no &  motion info. & a new feature by self-similarity of low-level features\\
MultiResC \cite{Park_MultiResC_ECCV_2010} & ECCV2010 & DPM & SW & HOG & SVM & NMS & yes&  yes &  ground plane & a multiresolution model based on DPM \& HOG\\
VeryFast \cite{Benenson_VeryFast_CVPR_2012} & CVPR2012 & CF & SW & ChnFtrs & boosting & NMS & yes & no  & geometric info. & very fast pedestrian detector running at 135 fps\\
CrossTalk \cite{Dollar_Crosstalk_ECCV_2012}  & ECCV2012 & CF & SW & ChnFtrs & boosting & NMS & no &  no & no & exploite local correlations for fast cascade design\\
MT-DPM \cite{Yan_MTDPM_CVPR_2013} & CVPR2013 & DPM & SW & HOG & SVM & NMS & no & yes  & ped./car relation &  mapping ped. of various scales to a common space\\
sDt \cite{Park_WS_CVPR_2013} & CVPR2013 & CF & SW & ChnFtrs & SVM & NMS & no & yes  & motion info. &  remove camera motion and object motion\\
SquaresChnFtrs \cite{Benenson_SquaresChnFtrs_CVPR_2013} & CVPR2013 & CF & SW & ChnFtrs & boosting & NMS & no&  no &  no & use square features to reduce randomness\\
Franken \cite{Mathias_Franken_ICCV_2013} & ICCV2013 & CF & SW & ChnFtrs & boosting & NMS & no &  yes & no & a fast training of many occlusion-specific classifiers\\
ACF \cite{Dollar_ACF_PAMI_2014} & PAMI2014 & CF & SW & ChnFtrs & boosting & NMS& no & no& no & aggregate local features by downsampling operation\\
InformedHaar \cite{Zhang_InformedHaar_CVPR_2014} & CVPR2014 & CF & SW & ChnFtrs & boosting & NMS& no & no &  no & local ternary features based on pedestrian shape\\
LDCF \cite{Nam_LDCF_NIPS_2014} & NIPS2014 & CF & SW & ChnFtrs & boosting & NMS& no & no & no & remove correlations in local neighborhoods\\
2Ped \cite{Ouyang_2PD_PAMI_2015} & PAMI2015 & DPM & SW & HOG & SVM & NMS & no & yes  & no &  spatial configuration
patterns of nearby pedestrians\\
FCF \cite{Zhang_FCF_CVPR_2015} & CVPR2015 & CF & SW & ChnFtrs & boosting & NMS & no& no& no & construct a filtered channel framework\\
SqatialPooling \cite{Paisitkriangkrai_SP_PAMI_2016} & PAMI2016 & CF & SW & ChnFtrs & boosting & NMS & no&  no &  no & extract the features based on spatial pooling\\
SCF \cite{Costea_SCF_CVPR_2016} & CVPR2016 & CF & SW & ChnFtrs & boosting & NMS& no & no& semantic seg. & add segmentation features as additional channels\\
NNNF \cite{Cao_NNNF_CVPR_2016} & CVPR2016 & CF & SW & ChnFtrs & boosting & NMS& no & no& no & non-neighbouring features based on inner attributes\\
\hline
\end{tabular}}
\label{table:benchmark_hf}
\end{table*}

Feature extraction is the key component in proposal generation and classification, where the aim is to represent the proposals with discriminative features. A variety of features, ranging from handcrafted \cite{Dollar_ICF_BMVC_2009,Dollar_ACF_PAMI_2014,Dalal_HOG_CVPR_2005} to   deep features \cite{Krizhevsky_ImageNet_NIPS_2012,Simonyan_VGG_arXiv_2014,He_ResNet_CVPR_2016,Jiang_CSN_TNNLS_2017}, are proposed. Based on the underlying feature extraction scheme, pedestrian detection approaches can be roughly divided into handcrafted features based approaches and deep features based approaches. Most handcrafted features are based on the operations of local difference or sum. One of the most popular handcrafted features is the histogram of oriented gradients (HOG) \cite{Dalal_HOG_CVPR_2005}, which captures the changes in local intensity. The fusion of HOG features with other visual cues, such as texture \cite{Yan_HOGLBP_ICCV_2010} and color \cite{Khan_CNHOG_CVPR_2012}, has also been investigated. Different to handcrafted features, deep features are typically extracted from the convolutional neural network (CNN). The CNNs learn invariant features through a series of convolution and pooling operations followed by one or more fully-connected (FC) layers. Features from deeper layers are discriminative, whereas the shallow layers contain low-level features with high spatial resolution. Fig. \ref{fig:feature} shows visualizations of both handcrafted and deep features on several example images. 

\begin{figure}[t]
\centering
\includegraphics[width=0.96\linewidth]{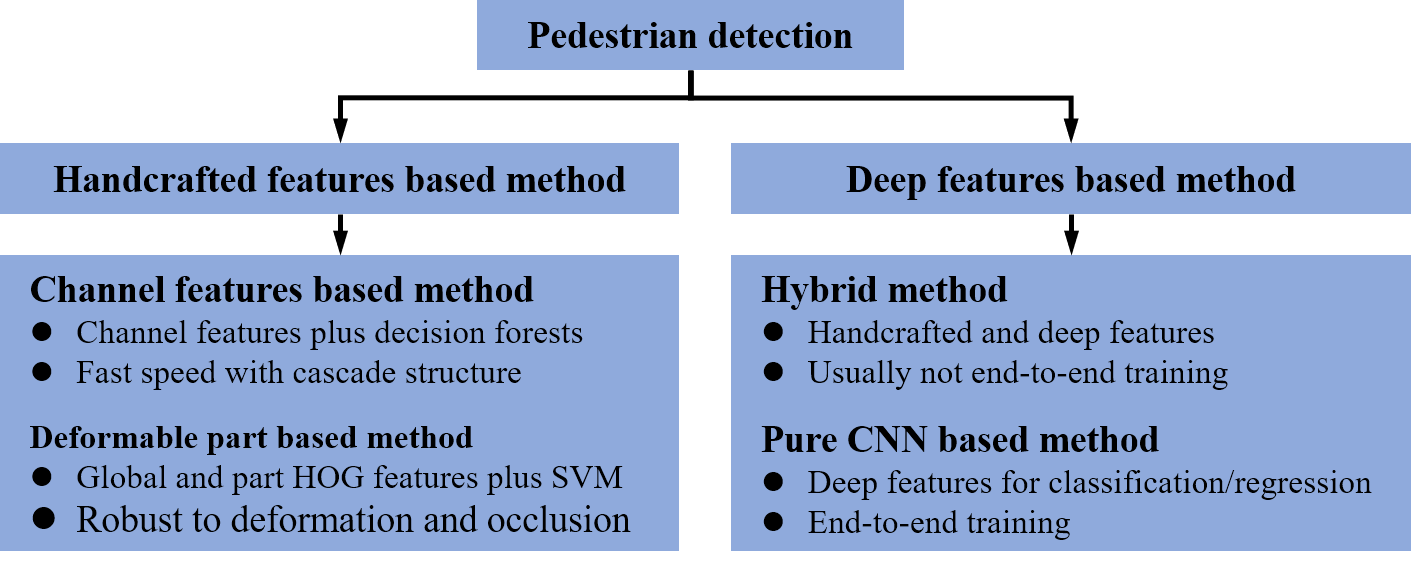}
\caption{Two different classes of single-spectral pedestrian detection approaches: handcrafted features based and deep features based methods. We further categorize the handcrafted based methods into channel features based and deformable part model based approaches. Further, deep features based pedestrian detection methods are categorized into hybrid and pure CNN based approaches. }
\label{fig:method1}
\end{figure}

\section{Single-spectral pedestrian detection}
\label{sec-sspd}
Most vision  applications, including pedestrian detection, acquire data using visible-light cameras since they are inexpensive and easily available. As such, most existing pedestrian detection methods \cite{Dollar_ICF_BMVC_2009,Benenson_TYPD_ECCV_2014,Dollar_Caltech_PAMI_2010,Zhang_Citypersons_CVPR_2017} employ this kind of data. We further separate (Fig. \ref{fig:method1}) these pedestrian detection methods into two main categories: handcrafted features based approaches and deep features based approaches. Moreover, the deep features based approaches are split into pure CNN based methods and hybrid methods. Next, we discuss the handcrafted features methods and then present a summary of deep features based methods.

\subsection{Handcrafted features based pedestrian detection}
Before the success of deep convolutional neural networks in computer vision tasks \cite{Krizhevsky_ImageNet_NIPS_2012,Girshick_RCNN_CVPR_2014,Ye_PurifyNet_TIFS_2019},
a variety of handcrafted feature descriptors, including SIFT \cite{Lowe_SIFT_IJCV_2004}, LBP \cite{Ojala_LBP_PAMI_2004}, SURF \cite{Bay_SURF_ECCV_2006}, HOG \cite{Dalal_HOG_CVPR_2005}, and Haar \cite{Viola_Haar_IJCV_2004}, have been investigated. These handcrafted features usually extract color, texture, or edge information. One of the most widely used handcrafted features for pedestrian detection \cite{Dalal_HOG_CVPR_2005} is histogram of oriented gradients (HOG). Further, most existing handcrafted based approaches either employ channel features \cite{Dollar_ACF_PAMI_2014,Nam_LDCF_NIPS_2014,Zhang_FCF_CVPR_2015} or deformable part models \cite{Yan_MTDPM_CVPR_2013,Yan_MTDPM_CVPR_2013,Felzenszwalb_CDPM_CVPR_2010} for the underlying model learning mechanism. Table \ref{table:benchmark_hf} summarizes some handcrafted features based methods. Before presenting a detailed introduction of these two kinds of approaches, we first describe their common inference and training steps.

\textit{Inference} Given an input image, handcrafted features are first extracted on different proposals (detection windows) generated by sliding the window with a fixed step (\textit{e.g.,} 2). Once the proposals are represented by handcrafted features, they are input to the trained pedestrian detector for prediction (classification). Since real-world pedestrians appear at different scales, input image is first resized at various scales and the detector is then applied on each scale to obtain predictions. Consequently, non-maximum suppression (NMS)  is utilized to remove duplicate bounding-boxes (proposals).

\textit{Training} The training proposals are generated by sliding-window methods. The proposals that have high overlap with ground-truths are treated as positive samples, otherwise they are treated as negative samples. Given positive and negative samples, the handcrafted features are extracted to represent these samples. Based on the extracted features, shallow classifiers (\textit{e.g.,} boosting or SVM) is used to learn a pedestrian detector to distinguish pedestrians (positive class) and the background (negative class). To improve detection performance, bootstrap technique \cite{Viola_Haar_IJCV_2004} is commonly adopted to select the hard samples over several training stages, where the hard negative samples at current stage are aggregated to the next one.  The detector trained after last stage is used during inference.

\begin{figure*}[t]
\centering
\includegraphics[width=0.99\linewidth]{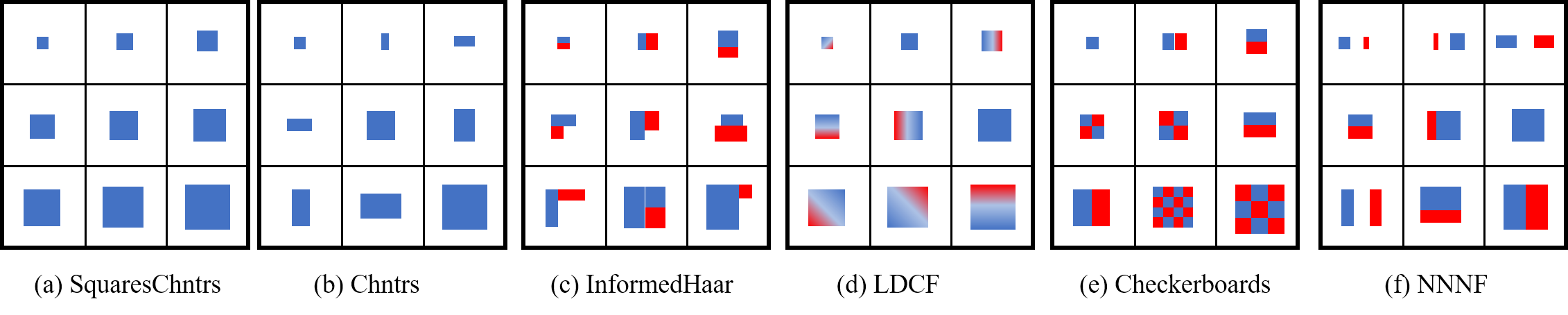}
\caption{Some typical features in channel features based methods, including SquaresChntrs \cite{Benenson_SquaresChnFtrs_CVPR_2013}, Chntrs \cite{Dollar_ICF_BMVC_2009}, InformedHaar \cite{Zhang_InformedHaar_CVPR_2014}, LDCF \cite{Nam_LDCF_NIPS_2014}, Checkererbords \cite{Zhang_FCF_CVPR_2015}, and NNNF \cite{Cao_NNNF_CVPR_2016}. The feature freedom degree in shape and space (local or non-local) becomes larger from left to right.} 
\label{fig:handfeatures}
\end{figure*}

\subsubsection{Channel features based methods}
Most channel features based methods extract a variety of local features from different types of channels (\textit{e.g.,} color and gradient channels) to represent each detection window (proposal) in an image. Then, they employ boosting technique together with decision forests to select a set of most discriminative features, which are used to train a pedestrian detector. One of the earlier detection methods belonging to this category is the popular Viola and Jones (VJ) detector \cite{Viola_Haar_IJCV_2004}. The VJ method first extracts the candidate Haar features for each proposal and then utilizes cascade AdaBoost \cite{Freund_AdaBoost_ML_1997} to learn the detector. Initially, the VJ method was for face detection.  However, compared to face detection, pedestrian detection is more challenging, and the initial VJ framework has been shown to be less effective on this task.

The seminal work of ChnFtrs \cite{Dollar_ICF_BMVC_2009} improves the VJ method for pedestrian detection. It first computes multiple registered image channels and then extracts the local sum features over these image channels. Afterwards, it utilizes the cascade AdaBoost to learn the detector. The ChnFtrs method shows that using ten registered channels (\textit{i.e.,} six gradient histograms, one gradient magnitude, and three LUV color channels) leads to state-of-the-art performance.

Based on the aforementioned registered channels (HOG+LUV) and boosting classifier, several variants of ChnFtrs have been proposed. Some methods \cite{Benenson_SquaresChnFtrs_CVPR_2013,Nam_LDCF_NIPS_2014,Dollar_ACF_PAMI_2014,Zhang_FCF_CVPR_2015} focus on extracting better local features from HOG+LUV channels. To reduce computational costs, Doll{\'a}r \textit{et al.} \cite{Dollar_ACF_PAMI_2014} proposed aggregated channel features (ACF) that aggregate feature values of every block as the candidate features. To avoid a large number of candidate features, Benenson \textit{et al.} \cite{Benenson_SquaresChnFtrs_CVPR_2013} selected the local sums of all the squares inside the
detection window as the candidate features. To remove the local correlations, Nam \textit{et al.} \cite{Nam_LDCF_NIPS_2014} and Zhou \textit{et al.} \cite{Zhou_GCSB_TITS_2019}  proposed to convolve the image channels with a fixed filter bank, learned from the training data, as the candidate features.  Zhang \textit{et al.} \cite{Zhang_FCF_CVPR_2015} built a generalized filtered channel framework, where several other detectors (\textit{e.g.,} ChnFtrs \cite{Dollar_ICF_BMVC_2009}, SquaresChnFtrs \cite{Benenson_SquaresChnFtrs_CVPR_2013}, and LDCF \cite{Nam_LDCF_NIPS_2014}) can be seen as the special cases. To avoid the large variety in the types of filter bank, Zhang \textit{et al.} \cite{Zhang_HowFar_CVPR_2016} further developed a small set of  filter banks inspired by LDCF. Shen \textit{et al.} \cite{Shen_LPDF_PR_2017,Shen_SPDF_TITS_2019,Shen_TAFT_TITS_2018} and Liu \textit{et al.} \cite{Liu_PDMP_TITS_2020} designed pixel neighborhood differential features for pedestrian detection. Li \textit{et al.} \cite{Li_LCOS_TITS_2016}  constructed the co-occurrence features in local neighborhoods using a binary pattern. 
Fu \textit{et al.} \cite{Fu_FSSS_ACCESS_2018} exploited the self-similar features based on linear discriminant analysis (LDA).  You \textit{et al.} \cite{You_ExtFCF_TITS_2018} proposed to use several convolutional layers to generate the channel features for pedestrian detection.


Besides the HOG+LUV channel features, other channel features have also been investigated. Costea \textit{et al.} \cite{Costea_SCF_CVPR_2016,Costea_MRFC_CVPR_2017} added semantic channel features as additional features. Paisitkriangkrai \textit{et al.} \cite{Paisitkriangkrai_SP_ECCV_2014,Paisitkriangkrai_SP_PAMI_2016} added the low-level visual features (\textit{i.e.,} covariance descriptor and LBP) and spatial pooling to the channel features.  Trichet and Bremond \cite{Trichet_LBPChannel_WACV_2018} introduced LBP-based channels to replace HOG+LUV channels.  Zhu \textit{et al.} \cite{Zhu_DLSF_Neurocomputing_2017} proposed additional high-level semantic features by using a sparse coding algorithm on mid-level image representations. In addition, some methods \cite{Walk_NF_CVPR_2010,Park_WS_CVPR_2013} use motion information to aid pedestrian detection. 

Since pedestrian detection only focuses on pedestrian category, some specific pedestrian characteristics can be exploited for feature design. Zhang \textit{et al.} \cite{Zhang_InformedHaar_CVPR_2014} incorporated the prior knowledge that pedestrians usually contain head, upper body, and lower body into Haar-like feature  design. Motivated by the human visual system, Zhang \textit{et al.} \cite{Zhang_HVDF_TCSVT_2015} further  developed the center-surround contrast features. Inspired by appearance constancy and shape symmetry of pedestrians, Cao \textit{et al.} \cite{Cao_NNNF_CVPR_2016}  designed two non-neighbouring features (\textit{i.e.,} side-inner difference features and shape symmetrical features). These non-neighbouring features have been shown to be complementary to the local features.

Fig. \ref{fig:handfeatures} shows some typical handcrafted channel features. The features contains local sum features, local difference features, haar features, non-neighbouring features, etc. From the left to the right, the features have a larger freedom degree in shape and space, and the corresponding methods have better performance in accuracy.

Most methods above strive for improved detection accuracy. In contrast, several other methods focus on improving speed. To reduce computational costs caused by image pyramid, Doll{\'a}r \textit{et al.} \cite{Dollar_FPPW_BMVC_2010} introduced fast feature pyramids, where the channel features at a single scale are used to approximate channel features at nearby scales. Further, Doll{\'a}r \textit{et al.} \cite{Dollar_Crosstalk_ECCV_2012} designed several fast cascade structures (\textit{i.e.,} soft cascade, excitatory cascade and inhibitory cascade).  Pang \textit{et al.} \cite{Pang_iPW_TCYB_2017} proposed to sample proposals in cascade stages according to the sampling distribution. Benenson \textit{et al.} \cite{Benenson_VeryFast_CVPR_2012} developed a fast pedestrian detector (called VeryFast) that trains multiple pedestrian detectors and shares the features for different detectors.  Rajaram \textit{et al.} \cite{Rajaram_ACFMR_TITS_2016} trained multiple multi-resolution ACF detectors for fast pedestrian detection.

\subsubsection{Deformable part based methods}
To better capture the deformation of objects such as pedestrians, the deformable part based model \cite{Felzenszwalb_DPM_PAMI_2010} (DPM) was introduced. DPM is one of the most popular handcrafted approaches for detecting both generic objects and pedestrians, which  consists of a coarse root model and a set of higher-resolution parts deformation models. The final score is equal to the score of the root model plus the sum over parts of the maximum of the part score minus a deformation cost. In each model, histograms of oriented gradients (HOG) \cite{Dalal_HOG_CVPR_2005} are used to extract the features. By dividing detection window into multiple spatial sub-regions (cells), the gradient histogram features are computed for each cell. Consequently, histograms of each cell are concatenated in a single feature representation to describe detection window. Since HOG features encode the variance in local shape (\textit{e.g.,} edge and gradient structure) very well and the part based model is able to capture the deformations, their combination in the deformable part based models yield promising results in 2006 PASCAL object detection challenge. 

\begin{table*}
\centering
\caption{Summary of 45 typical deep features based methods for pedestrian detection. These methods are typically built on convolutional neural networks. `P-CNN' means the pure CNN method, `Hybrid' means the hybrid method, and `SW' means the sliding-window strategy. `R-CNN' means the feature extraction fashion in R-CNN series, including R-CNN, Fast R-CNN, and Faster R-CNN.}
\resizebox{0.99\textwidth}{!}{
\setlength\tabcolsep{4pt}
\renewcommand\arraystretch{1.1}
\begin{tabular}{|l|l|c|c|c|c|c|c|c|c|l|}
\hline
Method & Publication & Family & Proposal & Feature & Classifier & Post-proc. & Scale-aware & Part-aware & Context & Description\\
\hline
\hline
DDN \cite{Ouyang_DDM_CVPR_2012} & CVPR2012 & Hybrid & SW & HOG & softmax & NMS& no & yes& no & first deep model for pedestrian detection\\
UMS \cite{Sermanet_UMSFL_CVPR_2013} & CVPR2013 & P-CNN & SW & CNN & softmax & NMS& no & no& ms fusion & one of the earliest deep pedestrian detectors\\
UDN \cite{Ouyang_UDN_ICCV_2013} & ICCV2013 & P-CNN & SW & CNN & softmax & NMS& no & yes& no & join different components by a deep network\\
SDN \cite{Luo_SDN_CVPR_2014} & CVPR2014 & Hybrid & HOG & CNN & boosting & NMS& no & yes& no & model mixture of visual variations by networks\\
ConvNet \cite{Hosang_convnet_CVPR_2015} & CVPR2015 & Hybrid & ACF & CNN & softmax & NMS& no & no& no & a state-of-the-art performance using convnets\\
TA-CNN \cite{Tian_TACNN_CVPR_2015} & CVPR2015 & Hybrid & ACF & CNN & boosting & NMS& no & no& attributes & join detection with multiple semantic tasks\\
DeepCascades \cite{Angelova1_DeepCascade_BMVC_2015} & BMVC2015 & Hybrid & VeryFast & CNN & softmax & NMS& no & no & no & one of first real-time and very accurate detector\\
CCF \cite{Yang_CCF_ICCV_2015} & ICCV2015 & Hybrid & ACF & CNN & boosting & NMS& no & no & no & extend FCF \cite{Zhang_FCF_CVPR_2015} to conv. channel features\\
DeepParts \cite{Tian_DeepParts_ICCV_2015} & ICCV2015 & Hybrid & ACF & CNN & SVM & NMS& no & yes& no & handle occlusion with deep part pool\\
CompACT-Deep \cite{Cai_CompACT_ICCV_2015} & ICCV2015 & Hybrid & SW & ChnFtrs+CNN & boosting & NMS& no & no & no & a complexity-aware cascade training structure\\
EEPD \cite{Stewart_EEP_CVPR_2016} & CVPR2016 & P-CNN & - & CNN & LSTM & no & no & no & no & end-to-end approach directly predicting objects\\
MS-CNN \cite{Cai_MSCNN_ECCV_2016} & ECCV2016 & P-CNN & RPN & FPN & softmax & NMS& yes & no & contextual RoI & multi-scale features for scale-ware detection\\
RPN+BF \cite{Zhang_RPN+BF_ECCV_2016} & ECCV2016 & Hybrid & RPN & CNN & softmax & NMS& no& no & no &  analyse limitations
of FR-CNN for pedestrians\\
MCF \cite{Cao_MCF_TIP_2016} & TIP2017 & Hybrid & SW &ChnFtrs+CNN & boosting & NMS& no & no& no & construct a multi-layer channel framework \\
SubCNN \cite{Xiang_SACNN_WACV_2017} & WACV2017 & P-CNN & RPN & R-CNN & softmax & NMS& no & no& no & joint detection and subcategory classification\\
F-DNN \cite{Du_FDNN_WACV_2017} & WACV2017 & P-CNN & SSD & CNNs & softmax & NMS& no & no& segmentation & a deep fusion of multiple networks\\
PGAN \cite{Li_PGAN_CVPR_2017} & CVPR2017 & P-CNN & RPN & R-CNN & softmax & NMS& yes & no& no & narrow feature differences by GAN\\
HyperLearner \cite{Mao_HyperLearner_CVPR_2017} & CVPR2017 & P-CNN & RPN & R-CNN & softmax & NMS& no & no & segmentation & learning extra features by multi-task learning\\
Adapted FR-CNN \cite{Zhang_Citypersons_CVPR_2017} & CVPR2017 & P-CNN & RPN & R-CNN & softmax & NMS& no & no& no & improved Faster R-CNN for pedestrians\\
JL-TopS \cite{Zhou_JL_ICCV_2017} & ICCV2017 & Hybrid & RPN & CNN & boosting & NMS& no & yes& no &  joint part detectors by multi-label learning \\
SDS-RCNN \cite{Brazil_SDS_ICCV_2017} & ICCV2017 & P-CNN & RPN & R-CNN & softmax & NMS& no & no& segmentation &  joint semantic segmentation and detection\\
PCN \cite{Wang_PCN_BMVC_2017} & BMVC2017 & P-CNN & RPN & R-CNN & softmax & NMS& no & yes& contextual RoI & use body parts semantic and context information\\
CFM \cite{Hu_CFM_CSVT_2018} & TCSVT2018 & Hybrid & SW & CNN & boosting & NMS& no& no & no & ensemble of boosted models by inner features\\
SAF-RCNN \cite{Li_SARCNN_TMM_2018} & TMM2018 & Hybrid & ACF & R-CNN & softmax & NMS& yes& no & no & two built-in sub-networks for different scales\\
SCNN \cite{Chen_SCNN_PAMI_2018} & PAMI2018 & Hybrid & ACF & CNN & softmax & NMS& no& no & no &  subcategory-aware network for intra-class variance\\
RepulsionLoss \cite{Wang_Repulsion_CVPR_2018} & CVPR2018 & P-CNN & RPN & R-CNN & softmax & NMS& no& no & no & novel repulsion loss for box regression\\
OHNH \cite{Noh_OHNH_CVPR_2018} & CVPR2018 & P-CNN & - & SSD & softmax & NMS & yes & no & no & part-aware score added in single-shot detector\\
FR-CNN ATT \cite{Zhang_FasterATT_CVPR_2018} & CVPR2018 & P-CNN & RPN & R-CNN & softmax & NMS& no& yes & no & channel attention mechanism for occlusion\\
Bi-Box \cite{Zhou_Bibox_ECCV_2018} & ECCV2018 & P-CNN & RPN & R-CNN & softmax & NMS& no& yes & no & two RoIs for fully/visible-body detections\\
GDFL \cite{Lin_GDFL_ECCV_2018} & ECCV2018 & P-CNN & - & SSD & softmax& NMS& yes& no  & ms fusion & encode fine-grained attention masks\\
OR-CNN \cite{Zhang_ORCNN_ECCV_2018} & ECCV2018 & P-CNN & RPN & R-CNN & softmax & NMS& no& yes & no & part occlusion-aware RoI pooling layer\\
TTL \cite{Song_TLL_ECCV_2018} & ECCV2018 & P-CNN & - & CNN & softmax & NMS& no& no & ms fusion &  use topological somatic line for detection\\
ALFNet \cite{Liu_ALFNet_ECCV_2018} & ECCV2018 & P-CNN & - & SSD & softmax& NMS & yes & no & no &  stack a series of predictors on SSD\\
CSP \cite{Liu_CSP_CVPR_2019} & CVPR2019 & P-CNN & - & CNN & softmax  & NMS& no& no & ms fusion &  one of first anchor-free pedestrian detector \\
Adaptive-NMS \cite{Liu_AdaptiveNMS_CVPR_2019} & CVPR2019 & P-CNN & -/RPN & SSD/FPN & softmax & Adapt. NMS& yes& no & no & dynamic NMS threshold based on target density\\
AR-Ped \cite{Brazil_ANP_CVPR_2019} & CVPR2019 & P-CNN & AR-RPN & FPN & softmax & NMS& yes& no & ms fusion & multi-phase autoregressive module for RPN\\
FRCN+A+DT \cite{Zhou_DFT_ICCV_2019} & ICCV2019 & P-CNN & RPN & R-CNN & softmax & NMS& no& no & no & narrow the occluded/unoccluded features\\
MGAN \cite{Pang_MGAN_ICCV_2019} & ICCV2019 & P-CNN & RPN & R-CNN & softmax & NMS& no& yes & no & mask-guided attention for RoI regions\\
PedHutter \cite{Chi_PedHunter_AAAI_2020} & AAAI2020 & P-CNN & RPN & FPN & softmax & NMS& yes& yes & no & mask-guided module encoding head information\\
JointDet \cite{Chi_JointDet_AAAI_2020} & AAAI2020 & P-CNN & RPN & R-CNN & softmax & RDM& no & yes & no &  head-body relationship discriminating module\\
PRNet \cite{Song_PRN_ECCV_2020} & ECCV2020 & P-CNN & - & SSD & softmax & NMS& no & yes & ms fusion &  a novel progressive refinement network\\
Case \cite{Xie_CSRCNN_ECCV_2020} & ECCV2020 & P-CNN & RPN & R-CNN & softmax & CaSe-NMS& no & no & no &  a count-weighted detection loss\\
PBM \cite{Huang_PBM_CVPR_2020} & CVPR2020 & P-CNN & RPN & R-CNN & softmax & R$^2$NMS& no & yes & no &  a novel NMS based on a paired-box model\\
TFAN \cite{Wu_TFAN_CVPR_2020} & CVPR2020 & P-CNN & RPN & R-CNN & softmax & NMS& no & no & no &   a tube feature aggregation network for occlusion\\
CrowdDetection \cite{Chu_CrowdDetection_CVPR_2020} & CVPR2020 & P-CNN & RPN & R-CNN & softmax & Set NMS& no & no & no &  predict multiple correlated instances per proposal\\
\hline
\end{tabular}}
\label{table:method_deepfeature}
\end{table*}

\begin{figure*}[t]
\centering
\includegraphics[width=0.99\linewidth]{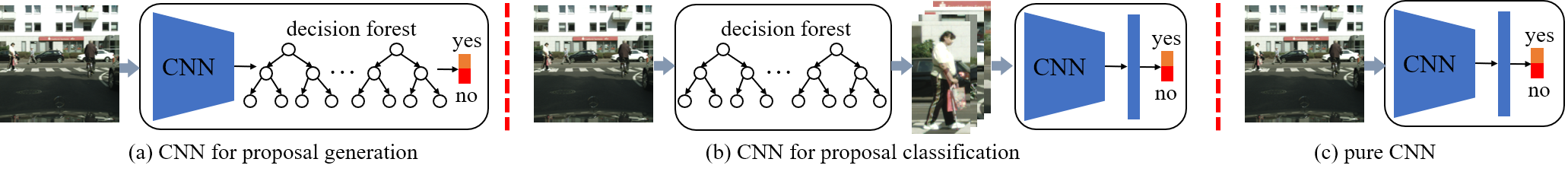}
\caption{Architectures of deep features based methods. (a) and (b) show two techniques in hybrid methods, while (c) is pure CNN based method. In (a), CNN extracts deep features for proposal generation and a shallow classifier is used for proposal classification. In (b), a handcrafted features based method is used for proposal generation and CNN is for proposal classification. In (c), pure CNN is used for both proposal generation and classification in an end-to-end fashion.}
\label{fig:cnn}
\end{figure*}

Several variants of the DPM model have been proposed in the literature \cite{Felzenszwalb_CDPM_CVPR_2010,Park_MultiResC_ECCV_2010,Yan_MTDPM_CVPR_2013,Jafari_TM_ICRA_2016}. Some of these variants focus on further improving pedestrian detection accuracy. Park \textit{et al.} \cite{Park_MultiResC_ECCV_2010} developed multi-resolution feature representations for pedestrians of various scales. Specifically, a deformable and high-resolution part-based model is used for large-sized pedestrian detection, while a rigid and low-resolution template is used for small-sized pedestrian detection. Yan \textit{et al.} \cite{Yan_MTDPM_CVPR_2013} proposed to map features from different resolutions to the same subspace using the proposed resolution-aware transformations based on the DPM detector. Ouyang \textit{et al.} \cite{Ouyang_2PD_CVPR_2013,Ouyang_2PD_PAMI_2015} improved single-pedestrian detection with the help of multi-pedestrian detection. The multi-pedestrian detector is learned by a mixture of deformable part-based models, where each single pedestrian is also treated as a part. Afterwards, the relationship between single-pedestrian detection  and multi-pedestrian detection is modelled to refine pedestrian detection performance. Wan \textit{et al.} \cite{Wang_SOA_ICIP_2016} incorporated  a scale prior and occlusion
analysis into deformable part models. Other than striving for improved accuracy, several works focus on improving detection speed. Felzenszwalb \textit{et al.} \cite{Felzenszwalb_CDPM_CVPR_2010}  proposed partial hypotheses to early reject some low scoring samples by using fewer models. As a result, the speed of DPM detector is 20 times faster, without sacrificing detection accuracy. Baek \textit{et al.} \cite{Baek_KP_TITS_2020} developed an additive kernel SVM and BING proposal generation method for fast pedestrian detection.

\subsection{Deep features based pedestrian detection}
In recent years, deep convolutional neural networks (CNN) have achieved great success in many computer vision tasks, (\textit{e.g.,} image classification \cite{Krizhevsky_ImageNet_NIPS_2012,He_ResNet_CVPR_2016,Huang_DenseNet_CVPR_2017}, semantic segmentation \cite{Long_FCN_CVPR_2017,Pang_BSG_ICCV_2019,Lu_SeeMore_CVPR_2019}, and object detection \cite{Girshick_FastRCNN_ICCV_2015,Redmon_YOLO_CVPR_2016,Ouyang_DeepID_PAMI_2017}). With the success of deep learning in generic object detection, several attempts have been made to apply deep CNN features to pedestrian detection \cite{Ouyang_DDM_CVPR_2012,Sermanet_UMSFL_CVPR_2013,Ouyang_UDN_ICCV_2013}. Table \ref{table:method_deepfeature} summarizes some deep features based pedestrian detection methods. In this sub-section, we split the related approaches into two categories: hybrid and pure CNN based methods.

\subsubsection{Hybrid pedestrian detection methods}
As in handcrafted approaches, hybrid methods also have proposal generation and classification steps. According to what CNN features are used for, we divide hybrid approaches into two classes. Some approaches employ CNN features for proposal generation and a shallow classifier for proposal classification (\textit{i.e., CNN for proposal generation} in Fig. \ref{fig:cnn}(a)), whereas some other methods use handcrafted methods for proposal generation and CNN features for proposal classification (\textit{i.e., CNN for proposal classification} in Fig. \ref{fig:cnn}(b)).  These hybrid approaches share  some  common  training  and  inference protocols, described next. Then,  we  present a discussion on different hybrid methods. 

\textit{Inference} (1) \textit{CNN for proposal generation}. Deep features are first extracted from the entire image. After that, the trained detector slides over the extracted feature map with a fixed step. At each position, the trained detector assigns detection window as either positive class (pedestrian) or negative class based on corresponding features. (2) \textit{CNN for proposal classification}. A handcrafted features based method is used to extract some candidate proposals. Then, the trained CNN classifier classifies these proposals into either the positive or negative class. For both these two kinds of methods, a non-maximum suppression technique is finally used to suppress the duplicate bounding-boxes.

\textit{Training} (1) \textit{CNN for proposal generation}. For positive and negative samples, deep features are extracted from pre-trained CNN. Based on these extracted features, a shallow classifier (\textit{e.g.,} boosting or SVM) along with the bootstrap technique is used to learn the pedestrian detector. The training samples are generated by sliding-window or handcrafted features based methods. (2) \textit{CNN for proposal classification}. First, the handcrafted features based methods are used to generate some candidate proposals. Based on these candidate proposals, a CNN with the softmax layer is trained in an end-to-end fashion (both proposal generation and classification) on the specific pedestrian dataset.

Some pedestrian detection approaches employ CNNs for proposal generation and a shallow classifier for proposal classification.  Yang \textit{et al.} \cite{Yang_CCF_ICCV_2015} proposed to replace the handcrafted filtered channel features (FCF) \cite{Zhang_FCF_CVPR_2015} with convolutional channel features (CCF), where each pixel in  the last convolutional layer is used as a single feature.  Hu \textit{et al.} \cite{Hu_CFM_CSVT_2018} trained an ensemble of boosted decision forests based on the features from the different layers of a CNN. Zhang \textit{et al.} \cite{Zhang_RPN+BF_ECCV_2016} and Tesemaa \textit{et al.} \cite{Tesema_HCPD_Neurocomputing_2020} utilized the region proposal network (RPN) as an initial pedestrian detector and further trained a shallow classifier with deep features to refine detection results. Li  \textit{et al.} \cite{Li_NF_Neurocomputing_2017} proposed to  extract multi-resolution deep features from different convolutional networks to learn a pedestrian detector. Sheng \textit{et al.} \cite{Sheng_FSDFC_Neurocomputing_2017} integrated deep semantic segmentation features and shallow handcrafted channel features into a filtered channel framework. Tesema \textit{et al.}  \cite{Tesema_HCPD_Neurocomputing_2020}  proposed to pool both handcrafted features and deep features to learn pedestrian detector with decision forests. Wang \textit{et al.} \cite{Wang_AEMSRPN_TITS_2019} developed a multi-scale  region proposal network to deal with a variance in scale and integrated a decision forest for classification.

Several other pedestrian detection works treat CNNs as a deep classifier to classify the candidate proposals. Hosang \textit{et al.} \cite{Hosang_convnet_CVPR_2015} provided a deep analysis on the effectiveness of CNNs for pedestrian detection. Based on their careful design, the simple CNNs were shown to achieve promising results for pedestrian detection. Tian \textit{et al.} \cite{Tian_DeepParts_ICCV_2015}   trained the multiple part detectors and then trained a linear SVM to combine the scores of part detectors.   Ribeiro \textit{et al.} \cite{Ribeiro_IPDCL_PR_2017} trained multiple deep networks with different inputs (\textit{e.g.,} color and segmentation images) to refine the results of ACF detector \cite{Dollar_ACF_PAMI_2014}. Ouyang \textit{et al.} \cite{Ouyang_JointDeep_ICCV_2013,Ouyang_UDN_PAMI_2018} built a unifying deep learning model to join different tasks (\textit{i.e.,} feature extraction, deformation handling, occlusion handling, and classification). Luo \textit{et al.} \cite{Luo_SDN_CVPR_2014} proposed to automatically learn hierarchical features, salience maps, and mixture representations of different body parts by a Switchable Restricted Boltzmann Machine. Jung \textit{et al.} \cite{Jung_DNGN_PRL_2017} developed a guiding network to assist the training of pedestrian detector. 

Besides the aforementioned approaches that strive for higher accuracy, other methods aim to improve detection speed. Cai \textit{et al.} \cite{Cai_CompACT_ICCV_2015} designed a complexity aware cascade strategy (CompACT) to balance accuracy and computational complexity. Specifically, CompACT uses the features of lower computational complexity at early stages and the features of higher computational complexity at later stages. Cao \textit{et al.} \cite{Cao_MCF_TIP_2016} designed multi-layer channel features (MCF), where the handcrafted channels and each layer of CNNs are integrated together. Based on the multi-layer feature channels, a multi-stage cascade detector is learned. MCF not only makes full use of features of different layers, but also efficiently rejects many samples at a lower computational cost.  Angelova \textit{et al.} \cite{Angelova1_DeepCascade_BMVC_2015} proposed to cascade the handcrafted detector (\textit{i.e.,} VeryFast \cite{Benenson_VeryFast_CVPR_2012}) and multiple deep networks for faster pedestrian detection. Jiang \textit{et al} \cite{Jiang_SUNN_Neurocomputing_2016} proposed to share deep features for multi-scale pedestrian detection.

\subsubsection{Pure CNN based pedestrian detection methods}
The success and popularity of Faster R-CNN \cite{Ren_FasterRCNN_NIPS_2015} for generic object detection prompted the construction of pure CNN based pedestrian detection approaches, where CNNs are used for both proposal generation and classification. Fig. \ref{fig:cnn}(c) shows the architecture of a pure CNN based pedestrian detector. Initially, the direct usage of Faster R-CNN for pedestrian detection resulted in below-expected performance. Zhang \textit{et al.} \cite{Zhang_Citypersons_CVPR_2017} introduced several modifications (\textit{e.g,} anchor scale and ignored region handling) to Faster R-CNN for improved pedestrian detection. Compared with hybrid methods, the pure CNN based approaches are more effective and simpler. Moreover, 
they are typically trained in an end-to-end fashion. We first describe  some  common training and inference protocols. Afterwards, we present  a  discussion  on  different  pure CNN based  methods.

\textit{Inference} Given a test image, deep features (of the entire image) are first extracted using a CNN. Some candidate proposals are first generated based on the default anchors and then classified by the corresponding features with a softmax layer and regressed to obtain a more accurate location. Consequently, a non-maximum suppression (NMS) technique is used to suppress duplicate bounding-boxes.

\textit{Training} Given a training image, deep features (of the entire image) are extracted using a CNN. The anchors are set as each position of feature maps and assigned as positive and negative samples. Based on the gradient back-propagation algorithm, CNNs are updated at each iteration. The learning rate, weight decay, and batch size are set according to the specific pedestrian detection dataset.

\begin{figure*}[t]
\centering
\includegraphics[width=0.99\linewidth]{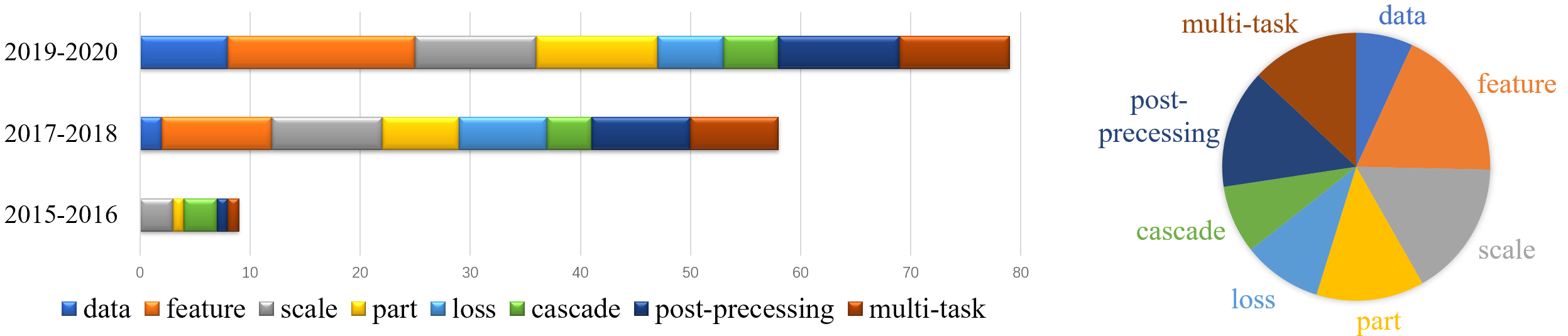}
\caption{Statistics of deep features based methods. The left part shows the change in the number of methods belonging different classes. Here, feature in legend indicates the union of feature-fused and attention-based methods which both aim to improve feature description ability. We ignore the anchor-free methods and others due to the limited number of these methods. The right part shows the percentage of different classes.}
\label{fig:deepmethods}
\end{figure*}

In recent years, a variety of pure CNN based pedestrian detection methods have been introduced in the literature. Next, we present a summary of these approaches.

\textbf{Scale-aware methods}  Scale-aware methods generally use the different in-network layers (or sub-networks) to detect objects at different scales. Some of these methods extract RoI features of proposals from different layers (\textit{called scale-aware RoI}). Yang \textit{et al.}  \cite{Yang_SDP_CVPR_2016} and Liu \textit{et al.} \cite{Liu_SAMRCNN_BMVC_2018} extracted the RoI features of proposals according to their scales. If the object has a smaller scale, the RoI features from the earlier layer are extracted. If the object has a larger scale, the RoI features from the later layers are extracted. Zhu \textit{et al.} \cite{Zhu_SADRN_ACCV_2016} flexibly chose the RoI features for regression and combined the RoI features from multiple layers for classification. SAF-RCNN \cite{Li_SARCNN_TMM_2018} integrates the scores of large-scale and small-scale sub-networks according to the proposal scales.
Some methods generate candidate proposals from different layers (\textit{called scale-aware RPN}). To make the receptive fields match the objects of different scales, Cai \textit{et al.} \cite{Cai_MSCNN_ECCV_2016} proposed to extract proposals from multiple in-network layers. Specifically, the lower layers with smaller receptive fields are employed for small objects, whereas the later layers with larger receptive fields are utilized for large objects. To enhance the feature semantics, Lin \textit{et al.}  \cite{Lin_FPN_CVPR_2017,Lin_FL_PAMI_2020} adopted a top-down structure to integrate the features from deep layer with the features from shallow layer (called FPN). Hu \textit{et al.} \cite{Hu_FPN++_ICME_2019} further modified FPN by reducing the convolutional stride from 2 to 1 at earlier layers to retain more information for small-scale pedestrian detection.

\textbf{Part-based methods} Features from the local part of an object play an important role in capturing occluded or deformable pedestrians. Several methods have investigated the integration of part-based information. Xu \textit{et al.} \cite{Xu_SPRCNN_CVPRW_2019} proposed to first detect the key-points of each proposal and then generate six parts based on these key-points. Afterwards, they combined the features of these parts together. Zhao \textit{et al.} \cite{Zhao_JHPCNN_BMVC_2018} introduced two branches for holistic  and part predictions and built a tree-structured module to integrate them. Zhang \textit{et al.} \cite{Zhang_ORCNN_ECCV_2018} proposed to combine the features of different parts for classification and regression.

Several recent pedestrian detection methods use the visible-body information of a pedestrian. Zhou \textit{et al.} \cite{Zhou_Bibox_ECCV_2018} trained a deep network with two output branches to detect full body and visible part. The results of two branches are fused to obtain improved pedestrian detection. Pang \textit{et al.} \cite{Pang_MGAN_ICCV_2019} developed a novel mask-guided attention network to enhance the features of visible regions. Some methods use the head information to aid pedestrian detection. Chi \textit{et al.} \cite{Chi_JointDet_AAAI_2020} designed a joint network for head and pedestrian detection with relationship discriminating module for detection in crowd.  Zhang \textit{et al.}  \cite{Zhang_DoubleAnchor_arXiv_2019} proposed double anchor region proposal networks to respectively detect human heads and bodies and a joint NMS to combine the detection results. Lin \textit{et al.} \cite{Lin_PedJointNet_ACCESS_2019} built two-branch networks for head-shoulder and full body predictions and introduced an adaptive fusion mechanism. Lu \textit{et al.} \cite{Lu_SHE_arXiv_2019} proposed semantic head detection in parallel with a body branch.

\textbf{Attention-based methods} These methods aim to enhance the features of pedestrians while suppressing the features of background. According to the underlying attention mechanism, we divide the related approaches into \textit{self-attention methods} and \textit{semantic-guided attention methods}. Self-attention methods use the attention mechanism to relate different positions of features. Zhang \textit{et al.} \cite{Zhang_FasterATT_CVPR_2018} observed that different channels represent different parts of an object and  utilized a channel-wise attention \cite{Hu_SENet_CVPR_2017} for occluded pedestrian detection. Zou \textit{et al.} \cite{Zou_AGNN_PRL_2020} proposed a spatial attention module to up-weight the features of visible part based on class activation mapping technique. Chen \textit{et al.} \cite{Chen_DFCA_ACCESS_2019} proposed the competitive attention to fuse the features from different convolutional layers. 

Semantic-guided attention approaches  aim at joining some high-level semantic task and pedestrian detection and can be also treated as multi-task methods.  Brazil \textit{et al.} \cite{Brazil_SDS_ICCV_2017} proposed a multi-task infusion framework for joint pedestrian detection and semantic segmentation for  both proposal generation and classification. Lin \textit{et al.} \cite{Lin_GDFL_ECCV_2018,Lin_MGDF_TCSVT_2019} designed a scale-aware attention module to make the detector better focus on the regions of pedestrians. Gajjar \textit{et al.} \cite{Gajjar_ViSHuD_CVPRW_2018} and Yun \textit{et al.} \cite{Yun_OPT_ACCESS_2019} proposed to use the visual saliency task as a pre-processing step to better focus on the regions of a pedestrian.

\textbf{Feature-fused methods} These methods aim to capture the useful contextual and semantic information by multi-scale feature fusion. Ren \textit{et al.} \cite{Ren_RRC_CVPR_2017} built a recurrent rolling convolution architecture to gradually aggregate contextual information from the different layers. Based on MS-CNN \cite{Cai_MSCNN_ECCV_2016}, Jung \textit{et al.} \cite{Jung_DMDnet_ICIP_2017} further combined the features of consecutive layers. In contrast, Chu \textit{et al.} \cite{Chu_MCFF_ACCESS_2018} combined the features from all different layers together to generate high-level features. Liu \textit{et al.} \cite{Liu_CoupleNet_arXiv_2019} proposed a gated feature extraction module by adaptively fusing multi-layer features. Shang \textit{et al.} \cite{Shang_ZoomNet_ACML_2018} introduced a complementary sub-network to generate the high-resolution feature map for small-scale object detection. Zhang \textit{et al.} \cite{Zhang_MFRCNN_TVT_2019} concatenated the RoI features from different layers along with the global context. Zhang \textit{et al.} \cite{Zhang_LADC_CVPR_2019} proposed a context feature embedding with a standard convolution and a deformable convolution. Fei \textit{et al.} \cite{Fei_CAFL_ACCESS_2019} developed a new pixel-level context embedding module by integrating multi-cue context into a deep CNN feature hierarchy. Wang \textit{et al.} \cite{Wang_BPSCI_TMM_2018} proposed a local competition mechanism (maxout) for adaptive context fusion. Cao \textit{et al.} \cite{Cao_LBST_TIP_2020} embedded the large-kernel convolution into feature pyramid structure to exploit contextual information. Wu \textit{et al.} \cite{Wu_TFAN_CVPR_2020} proposed to adaptively fuse the features of current frame and  nearby frames. 

\textbf{Cascade-based methods} To improve localization quality, cascade structure has been widely used in generic object detection \cite{Cai_Cascade_CVPR_2018,Cai_Cascade_PAMI_2019,Zhang_RefineDet_CVPR_2018}. Recently, some methods have adopted cascade structure for pedestrian detection. Liu \textit{et al.} \cite{Liu_ALFNet_ECCV_2018} stacked multiple head predictors for multi-stage regressions. Brazil \textit{et al.} \cite{Brazil_ANP_CVPR_2019} designed  a multi-phase auto-regressive module, where each module is trained using increasingly precise labeling policies. Zhang \textit{et al.} \cite{Zhang_CircleNet_TITS_2019} proposed to detect the low resolution and occluded objects again at a finer scale by mimicking the process of humans. Ujjwal \textit{et al.} \cite{Ujjwal_OHS_WACV_2020}  first utilized semantic segmentation to select a small set of anchors and then re-pooled the features for classification and regression. Du \textit{et al.} \cite{Du_FDNN_WACV_2017} proposed to fuse the detection scores of multiple networks by a cascade soft-region rejection strategy. Hasan \textit{et al.} \cite{Hasan_EIR_arXiv_2020} combined Cascade R-CNN \cite{Cai_Cascade_CVPR_2018} and HRNet \cite{Wang_HRNet_PAMI_2020}  to achieve improved pedestrian detection performance. Song \textit{et al.} \cite{Song_PRN_ECCV_2020} proposed to divide pedestrian detection into three-phase steps: visible part prediction, anchor calibration, and full-body prediction.

\textbf{Anchor-free methods} These methods directly predict the score and pedestrian location or shape at each position. Compared with the anchor-based methods, the anchor-free methods avoid the handcrafted design with respect to the scale and aspect ratio of anchors, thereby having a simpler design and good generalization ability on different datasets. Song \textit{et al.} \cite{Song_TLL_ECCV_2018} propose to localize pedestrians by the somatic topological line.  Liu \textit{et al.} \cite{Liu_CSP_CVPR_2019} proposed to  predict the center point and the height of the pedestrian based on the high-level semantic feature maps. Zhang \textit{et al.}  \cite{Zhang_APD_arXiv_2019} treated each positive instance as a feature vector to encode both density and diversity information simultaneously.

Instead of focusing on the network design, some methods, discussed next, focus on data augmentation, loss learning, post processing, and multi-task learning.

\textbf{Data-augmentation based methods} These methods aim to improve detection performance. Some methods focus on generating more pedestrians or images (\textit{data generation}). Based on the prior knowledge of camera parameters, Hattori \textit{et al.} \cite{Hattori_SLSV_IJCV_2018} generated a variety of geometrically accurate images of synthetic pedestrians. Vobecky \textit{et al.} \cite{Vobecky_APDA_ICCVW_2019} used GAN \cite{Goodfellow_GAN_NIPS_2014} to generate people images in a required pose. Wu \textit{et al.} \cite{Wu_PMCGAN_BMVC_2019} developed a multi-modal cascaded generative adversarial network with U-net structure  to generate pedestrian data. Chen \textit{et al.} \cite{Chen_STDA_arXiv_2019} transformed real pedestrians from the same dataset into different shapes using the shape-guided deformation. Some methods focus on making full use of current data (\textit{data processing}). To improve occluded pedestrian detection, Chi \textit{et al.} \cite{Chi_PedHunter_AAAI_2020} added some occlusions to pedestrians. To generate better positive samples, Lu \textit{et al.} selected samples based on visible intersection-over-union. Zhao \textit{et al.} \cite{Zhao_CCCNN_TITS_2019}  introduced a strict matching metric for training sample generation by considering the alignments of different parts simultaneously.  Wei \textit{et al.} \cite{Wei_EOD_TITS_2019} proposed to use soft-NMS \cite{Bodla_SoftNMS_ICCV_2017} to select some occluded samples for training. Luo \textit{et al.} \cite{Luo_W3Net_CVPR_2020} proposed to use multi-modal data, including bird view
map, depth and corpus information, for pedestrian localization, scale prediction and classification.

\begin{table*}
\centering
\caption{Summary of 12 typical methods for multispectral pedestrian detection. `P-CNN' means the pure CNN method, `Hybrid' means the hybrid method.}
\resizebox{0.99\textwidth}{!}{
\setlength\tabcolsep{4pt}
\renewcommand\arraystretch{1.1}
\begin{tabular}{|l|l|c|c|c|c|c|c|c|c|l|}
\hline
Method & Publication & Family & Proposal & Feature & Classifier & Post-proc. & Scale-aware & Part-aware & Context & Description\\
\hline
\hline
ACF-C-T \cite{Hwang_KAIST_CVPR_2015} & CVPR2015 & CF & SW & ChnFtrs & boosting & NMS& no& no & no & extended ACF with the thermal channel\\
Halfway \cite{Liu_MPNN_BMVC_2016} & BMVC2016 & P-CNN & RPN & R-CNN & softmax & NMS& no& no & no & fused channel features at middle-level layers\\
CMT-CNN \cite{Xu_CMTCNN_CVPR_2017} & CVPR2017 & Hybrid & ACF & R-CNN & softmax & NMS& no& no & no & cross-domain features by cross-modality learning\\
MRFC \cite{Costea_MRFC_CVPR_2017} & CVPR2017 & CF & SW & ChnFtrs & boosting & NMS& no& no & 2D/3D &  multimodal multiresolution channel features\\
Fusion RPN \cite{Konig_FusionRPN_CVPR_2017} & CVPRW2017 & Hybrid & RPN & CNN & boosting & NMS& no& no & no & use pre-trained convnet by network in network\\
APF \cite{Park_APF_PR_2018} & PR2018 & P-CNN & RPN & R-CNN & softmax & NMS& no& no & no & channel weighting \& probability fusion\\
MSDS-RCNN \cite{Li_MSDS_BMVC_2018} & BMVC2018 & P-CNN & RPN & R-CNN & softmax & NMS& no& no & segmentation & joint detection and semantic segmentation tasks\\
TS-RPN \cite{Cao_TSRPN_IF_2019} & IF2019 & P-CNN & TS-RPN & CNN & softmax & NMS& no& no & no & adapt visible detector to multispectral domain\\
IAFR-CNN \cite{Li_IAFRCNN_PR_2019} & PR2019 & P-CNN & RPN & R-CNN & softmax & NMS& no& no & no & adaptively merge results by illumination value\\
HMFFN \cite{Cao_HMFFN_ISPRS_2019} & ISPRS2019 & P-CNN & RPN & R-CNN & softmax & NMS& no& no & no & box-level
segmentation supervised learning\\
AR-CNN \cite{Zhang_ARRCNN_ICCV_2019} & ICCV2019 & P-CNN & RPN & R-CNN & softmax & NMS& no& no & Contextual RoI & first work that tackles position shift problem\\
MBNet \cite{Zhou_AMIP_ECCV_2020} & ECCV2020 & P-CNN & - & SSD & softmax & NMS& yes& no & no& designing a modality balance network\\
\hline
\end{tabular}}
\label{table:method_ms}
\end{table*}

\textbf{Loss-driven methods} These methods either use new functions or add extra loss functions for pedestrian detection. Wang \textit{et al.} \cite{Wang_Repulsion_CVPR_2018} proposed two types of repulsion loss (\textit{i.e.,} RepGT loss and RepBox loss) for crowded pedestrian detection. The RepGT loss penalizes the predicted bounding-box  near other objects, whereas the RepBox loss makes the predicted bounding-box farther away from other predicted bounding-boxes, in case of belonging to different objects. Wu \textit{et al.}  \cite{Wu_BCNN_WACV_2017} developed a weighted loss that emphasizes challenging samples. Xiang \textit{et al.} \cite{Xiang_SACNN_WACV_2017} explicitly employed the loss of sub-category classification for pedestrian detection. Some methods use the loss function to narrow the feature gap between different samples.  Li \textit{et al.} \cite{Li_PGAN_CVPR_2017} developed an architecture that internally lifts representations of small objects to that of large objects. Zhou \textit{et al.} \cite{Zhou_DFT_ICCV_2019} proposed a discriminative feature transformation to make the pedestrian features approach the feature center of non-occluded pedestrians. Li \textit{et al.} \cite{Li_Mimic_CVPR_2017} proposed to narrow the feature gap between the small network and  the large network for efficient detection. Chen \textit{et al.} \cite{Chen_PDHKD_ICIP_2019} performed multi-stage distillation to learn the light-weight network for acceleration. Li \textit{et al.} \cite{Li_FDML_ICIP_2018} transformed the LR feature space into a new LR classification space using an optimal Mahanalobis metric. Xie \textit{et al.} \cite{Xie_CSRCNN_ECCV_2020} proposed to assign a large weight to the proposal in crowded scene.

\textbf{Post-processing methods} Some methods improve NMS to better combine detection results. Liu \textit{et al.} \cite{Liu_AdaptiveNMS_CVPR_2019} applied a dynamic suppression threshold based on the target density. Yang \textit{et al.} \cite{Yang_SGNMS_arXiv_2019} developed bounding-box-level Semantics-Geometry Embedding (SGE) to distinguish two heavily-overlapping boxes. Huang \textit{et al.} \cite{Huang_PBM_CVPR_2020} proposed R$^2$NMS that uses the IoU between
visible regions to determine
whether or not two full-body boxes overlap. Stewart \textit{et al.} \cite{Stewart_EEP_CVPR_2016} built an end-to-end network to directly predict the objects without post-processing. Wang \textit{et al.} \cite{Wang_S3D_CSVT_2020} proposed to set the confidence threshold by investigating the relationship between the scores and scales of pedestrians. Zhang \textit{et al.} \cite{Zhang_LED_ICIP_2018} designed an accurate localization-quality estimation module to refine classification scores. Yang \textit{et al.} \cite{Yang_KF_ACCESS_2019} developed a Kalman filter-based convolutional network to remove some false positives for pedestrian detection in videos.

\textbf{Multi-task methods} Some methods utilize semantic information to aid pedestrian detection. Mao \textit{et al.} \cite{Mao_HyperLearner_CVPR_2017} investigated the impact of aggregating additional features (\textit{e.g.,} segmentation, heatmap, disparity, and optical flow) for pedestrian detection by using a multi-task learning network. Wang \textit{et al.} \cite{Wang_DLSM_TCSVT_2019} proposed joint  semantic segmentation and pedestrian detection to better distinguish the background and foreground. Kishore \textit{et al.} \cite{Kishore_ClueNet_BMVC_2019} and  Zhao \textit{et al.} \cite{Zhao_HPR_TIP_2019} proposed to join occluded pedestrian detection and pose estimation in cascade structure. Han \textit{et al.} \cite{Han_QGPR_CVPR_2019} proposed to join pedestrian detection and person search.

\textbf{Others} Most methods above focus on pedestrian detection in color images. Recently, some methods have been proposed for thermal images or fish-eye images. Guo \textit{et al.} \cite{Guo_DAPD_ICIP_2019} designed a domain adaptation component to use the abundant color images associated with pedestrian annotations in thermal domain. Ghose \textit{et al.} \cite{Ghose_TISM_CVPRW_2019} proposed to use saliency to augment pedestrian detector in the thermal domain. Kieu \textit{et al.} \cite{Kieu_TCDA_ECCV_2020} designed a task-conditioned architecture to adapt pedestrian detector to the thermal domain. Qian \textit{et al.} \cite{Qian_OSTN_TMM_2020} introduced a projective model to transform normal images into fish-eye images and designed an oriented spatial transformer network to rectify warped pedestrian features for better recognition. 
Peng \textit{et al.} \cite{Peng_SFL_WACV_2019} proposed a new cost function for training object detectors on fish-eye images. 
Li \textit{et al.} \cite{Li_PDHW_TIE_2019} proposed to use the depth-wise separable convolution, linear bottleneck, and multi-scale feature fusion for pedestrian detection in hazy weather. To avoid annotating a large number of pedestrians, Wu \textit{et al.} \cite{Wu_ETD_TIP_2018,Wu_RPNV_TIP_2020}  developed a semi-supervised approach to train deep convolutional networks on partially labeled data. 

Fig. \ref{fig:deepmethods} provides the statistics of deep features based methods from 2015 to 2020. The left part shows the change in the number of methods belonging to different classes. It can be seen that data-augmentation based methods, feature enhanced methods, and part-based methods have a large increment in past two years. The right part shows the percentage of all the methods over the past six years. The feature-enhanced, post-processing, scale-aware, part-based methods are the dominant approaches. Among these methods, post-processing and part methods usually focus on occluded pedestrian detection, while scale-aware and feature-enhanced methods mainly deal with scale variance problem. Thus, most recent methods still focus on solving the problems of occlusion and scale-variance.

\begin{table*}
\centering
\caption{Summary of pedestrian datasets. The top is early pedestrian datasets, the middle is modern pedestrian datasets, and the bottom is multispectral pedestrian datasets.`full' means fully-body bounding-box, `visible' means visible-body bounding-box, and `head' means head bounding-box.}
\resizebox{0.99\textwidth}{!}{
\setlength\tabcolsep{4pt}
\renewcommand\arraystretch{1.1}
\begin{tabular}{|l|l|c|c|c|c|c|l|}
\hline
Name & Publication & \#Images & \#Pedestrians & Resolutions & Annotations & Time &Description \\
\hline
\hline
MIT \cite{Papageorgiou_MIT_IJCV_2000} & IJCV2000 & - & 924 & 64$\times$128 & full & day& one of earliest pedestrian datasets\\
INRIA \cite{Dalal_HOG_CVPR_2005}  & CVPR2005 & 2120 & 1774 & 640$\times$480 & full & day& one of earliest popular pedestrian datasets\\
ETH \cite{Ess_ETH_ICCV_2007} & ICCV2007 & 1803 & 12$k$ & 640$\times$480 & full & day & a pair of images in busy shopping streets\\
TUD-Brussels \cite{Wojek_TUD_CVPR_2009}  & CVPR2009 & 508 & 1326 & 640$\times$480 & full & day& pedestrians in the inner city of Brussels\\
Daimler \cite{Enzweiler_Daimler_PAMI_2009} & PAMI2009 & 29$k$ & 72$k$ & 640$\times$480 & full & day& gray-color images in urban traffic\\
\hline
\hline
Caltech \cite{Dollar_Caltech_PAMI_2010} & PAMI2010 & 250$k$ & 289$k$ & 640$\times$480& full, visible & day& a standard and complete pedestrian datasts\\
KITTI \cite{Geiger_KITTI_CVPR_2012}  & CVPR2012 & 15$k$ & 9$k$ & 1240$\times$376& full & day & a real-world computer vision benchmarks\\
CityPersons \cite{Zhang_Citypersons_CVPR_2017}  & CVPR2017 & 5$k$ & 32$k$ & 2048$\times$1024& full, visible & day& extensions on top of the Cityscapes \cite{Cordts_Cityscapes_CVPR_2016}\\
CrowdHuman \cite{Shao_CrowdHuman_arXiv_2018}  & arXiv2018 & 24$k$ & 552$k$ & -& full, visible, head & day& humans in crowded scenes from website\\
EuroCity \cite{Braun_EuroCity_PAMI_2019}  & PAMI2019 & 47$k$ & 219$k$  & 1920$\times$1024& full & day, night & images in multiple European Cities\\
NightOwls \cite{Neumann_NightOwls_ACCV_2018} & ACCV2019 & 281$k$  & 56$k$ & 1024$\times$640& full & night & pedestrians at night in three countries\\
WIDER Pedestrian  & Challenge & 97$k$  & 307$k$ & - & full & day & pedestrians in traffic and surveillance scenes\\
WiderPerson \cite{Zhang_WiderPerson_TMM_2020}  & TMM2019 & 13$k$  & 39$k$ & - & full & day & persons in the wild, not only traffic\\
TJU-Pedestrian \cite{Pang_DHD_TIP_2020}  & TIP2020 & 75$k$  & 373$k$ & - & full, visible & day, night & a diverse dataset in traffic and campus\\
\hline
\hline
KAIST \cite{Hwang_KAIST_CVPR_2015} & CVPR2015 & 95$k$ & 103$k$ & 640$\times$480& full & day, night & color-thermal image pairs in traffic scene\\
CVC-14 \cite{Gonzalez_CVC_Sensors_2016} & Sensors2016 & 5051 & 7795 & 640$\times$512& full & day, night & multimodal (FIR+visible) videosequences\\
\hline
\end{tabular}}
\label{table:datasets}
\end{table*}

\section{Multispectral pedestrian detection}
\label{sec-mspd}
In Section \ref{sec-sspd}, most reviewed methods focus on detecting pedestrians in single-spectral images (\textit{e.g.,} color image). However, single-spectral pedestrian detection is not very robust to illumination variations. For instance, the color camera is ineffective at acquiring useful information  at night. Therefore, multispectral pedestrian detection \cite{Chen_PDMSC_TIV_2019} becomes important for self-driving and video surveillance, where the color and thermal images provide complementary visual information. Table \ref{table:method_ms} summarizes some typical methods for multispectral pedestrian detection.

Some methods explore how to deep fusion of  multispectral images, including  input fusion, feature fusion, and decision fusion. Liu \textit{et al.} \cite{Liu_MPNN_BMVC_2016} exploited  four kinds of fusions at different stages (called low-level fusion, middle-level fusion, high-level fusion, and score fusion). It is found that middle-level fusion (Highway Fusion) achieves the best detection performance. To take advantage of the pre-trained model on ImageNet \cite{Russakovsky_ImageNet_IJCV_2015},  Konig \textit{et al.} \cite{Konig_FusionRPN_CVPR_2017} added one 1$\times1$ convolutional layer to reduce the number of fused channels to the same number of input channel of VGG. Further, boosted decision trees were used  as in \cite{Zhang_RPN+BF_ECCV_2016}. 
To better combine the features from different modalities, Zhang \textit{et al.} \cite{Zhang_CMIAN_IF_2019} introduced a cross-modality interactive attention module to exploit the complementary nature of different modalities. Guan \textit{et al.} \cite{Guan_IADNN_IF_2019} and Li \textit{et al.} \cite{Li_IAFRCNN_PR_2019}  explored an illumination-aware mechanism by using the predicted illumination value to re-weight the results of the day and night sub-networks. To solve the modality
imbalance problem, Zhou \textit{et al.} \cite{Zhou_MBNet_ECCV_2020} proposed a single-stage detector that contains a differential modality aware fusion module and an illumination aware feature alignment module.

Besides,  to solve the position mismatch problem between color image and thermal image, Zhang \textit{et al.} \cite{Zhang_ARRCNN_ICCV_2019} proposed  to capture the position shift and align the region features. Based on the aligned features, a multimodal re-weighted module was further introduced to generate reliable features. To generate more diversified proposals and learn better features, Li \textit{et al.} \cite{Li_MSDS_BMVC_2018} added two head-networks for joint semantic segmentation and pedestrian detection to the color image branch and thermal image branch during training.

For unsupervised domain adaptation, Guan \textit{et al.} \cite{Guan_UDA_CVPRW_2019} proposed to iteratively generate training labels and update the network parameters in the target domain. To automatically transfer a detector from a visible domain to a new multispectral domain without manual annotations, Cao \textit{et al.} \cite{Cao_TSRPN_IF_2019} presented an auto-annotation framework to label pedestrian instances in visible and thermal channels by leveraging the complementary information of multispectral data. Xu \textit{et al.} \cite{Xu_CMTCNN_CVPR_2017} designed a cross-modality learning framework to model the relations between color image and infrared image. Afterwards, features extracted from the cross-modality network are fused with the features extracted from the traditional detection network.

\section{Dataset and evaluation}
\label{sec-data}
Pedestrian datasets can be split into two classes: earlier pedestrian datasets and modern pedestrian datasets. Additionally, some multispectral pedestrian datasets have been also built. We first provide an introduction to these datasets. Then, we discuss different evaluation metrics. Finally we give a detailed comparison and analysis on these datasets.

\subsection{Earlier pedestrian datasets}
The earlier datasets in Table \ref{table:datasets} (top) are relatively small and mainly used by handcrafted features based methods.

\textbf{MIT} \cite{Papageorgiou_MIT_IJCV_2000} is one of the first pedestrian datasets. The training set contains 924 positive samples and 11,361 negative samples, where the resolution of samples is  of 128$\times$64 pixels. There are 123 images for testing in this dataset.

\textbf{INRIA} \cite{Dalal_HOG_CVPR_2005} is one of the most popular pedestrian datasets for handcrafted features based methods. This dataset mainly contains personal digital images. The training set contains 614 positive images with 1,208 pedestrians and 1,218 negative images. The test set has 288 images.

textbf{ETH} \cite{Ess_ETH_ICCV_2007} is taken on different days in busy shopping streets. The dataset contains three subsets for evaluation (\textit{i.e.,} `BAHNHOF', `JELMOLI', and `SUNNY DAY'). The `BAHNHOF' set has 999 images, the `JELMOLI' set has 450 images, and the `SUNNY DAY' has 354 images.

\textbf{TUD-Brussels} \cite{Wojek_TUD_CVPR_2009} is recorded from a driving car in the city of Brussels and contains 508 images at a fixed resolution of 650$\times$480 pixels. Additionally, this dataset contains temporal image pairs to learn motion features.

\textbf{Daimler} \cite{Enzweiler_Daimler_PAMI_2009} is a large-scale pedestrian dataset recorded at various times of day. There are 6,755 images for training and 21,790 images for testing. Different from other datasets, this dataset consists of gray images instead of color images.

\subsection{Modern pedestrian datasets}
In the recent era of deep learning, performance on  the early datasets has become saturated. As such, the research community has dedicated significant efforts towards building new modern pedestrian datasets. Modern pedestrian datasets in Table \ref{table:datasets} (medium) are significantly larger and aim for a more standard evaluation. Specifically, the number of images and pedestrians is usually over 10 times larger, and  a more unifying training and test data are provided.

\textbf{Caltech}\footnote{\url{www.vision.caltech.edu/Image_Datasets/CaltechPedestrians}} \cite{Dollar_Caltech_PAMI_2010} is one of most complete benchmarks for pedestrian detection. This dataset contains 11 video sets taken from the urban, where the first 6 video sets are used for training and the remaining 5 video sets are used for testing. Generally, the training images are captured by every 3\textit{rd} frame and the test images are captured by every 30\textit{th} frame.  In addition to the full-body bounding-box annotations, the visible-body bounding-box annotations are also provided.

\textbf{KITTI}\footnote{\url{www.cvlibs.net/datasets/kitti}} \cite{Geiger_KITTI_CVPR_2012} is a challenging computer vision benchmark, including multiple different vision tasks. For object detection (car detection, pedestrian detection, and cyclist detection), there are 7,481 training images and 7,518 test images. The resolution of images is about 1,240 $\times$ 376 pixels.

\textbf{CityPersons}\footnote{\url{https://github.com/cvgroup-njust/CityPersons}} \cite{Zhang_Citypersons_CVPR_2017} is a diverse pedestrian dataset built on the Cityscapes dataset \cite{Cordts_Cityscapes_CVPR_2016}. There are 2,975 training images, 500 validation images, and 1,575 test images. Compared with the early pedestrian dataset, CityPersons dataset is richer in diversity (\textit{i.e.,} different cities, different seasons, various weather conditions, and more persons per image).

\textbf{CrowdHuman}\footnote{\url{https://www.crowdhuman.org/}} \cite{Shao_CrowdHuman_arXiv_2018} is a recently collected dataset for crowded human detection. It contains 15,000 training images, 4,370 validation images, and 5,000 test images. There are about 23 persons per image. The bounding-box annotations of full-body, visible-body , and head are provided.

\textbf{EuroCity persons}\footnote{\url{https://eurocity-dataset.tudelft.nl}} \cite{Braun_EuroCity_PAMI_2019} is a large-scale dataset of urban scenes captured at both daytime and nighttime in multiple European cites. It contains three different person categories (\textit{i.e.,}  pedestrians, cyclists, and other riders). There are 47,337 images in total at a resolution of 1,920$\times$1,080 pixels.

\textbf{NightOwls}\footnote{\url{https://www.nightowls-dataset.org}} \cite{Neumann_NightOwls_ACCV_2018} is a pedestrian dataset recorded at nighttime, which aims to promote progress in pedestrian detection at night. There are about 128$k$ training images, 51$k$ validation images, and 103$k$ test images.

\textbf{WIDER Pedestrians}\footnote{\url{https://wider-challenge.org/}} focuses on detecting pedestrians and cyclists for surveillance and car-driving. It contains 96,500 images with 307,183 annotations. There are 8,240 images (58,190 annotations) in surveillance scenes  and 88,260  images (248,993 annotations) in car-driving scenes.

\textbf{WiderPerson}\footnote{\url{http://www.cbsr.ia.ac.cn/users/sfzhang/WiderPerson}} \cite{Zhang_WiderPerson_TMM_2020} focuses on pedestrian detection in the wild under multiple different scenes and is not limited to traffic scenes. The dataset contains 13,382 images with 400$k$ annotations with various kinds of occlusions, which are collected from the website using 50 keywords.

\textbf{TJU-Pedestrian}\footnote{\url{https://github.com/tjubiit/TJU-DHD}} \cite{Pang_DHD_TIP_2020} is a diverse high-resolution dataset, including two sets of TJU-Pedestrian-Traffic and TJU-Pedestrian-Campus. TJU-Pedestrian-Traffic has 20,388 images with 43,618 annotations, while TJU-Pedestrian-Campus has 55,088 images with 329,623 annotations.

\begin{algorithm}[t] 
\caption{Greedy matching strategy to compute true positives and false positives.} 
\label{Alg:GreedyMatch} 
\begin{algorithmic}[1] 
\REQUIRE ~~\\
The set of detection results $B_d$; 
The set of detection scores $S_d$;
The set of ground-truths $B_g$;
The confidence threshold $\alpha$;\\
\ENSURE ~~\\
The set of true positives $B_{tp}$;
The set of false positives $B_{fp}$;
The set of false negatives $B_{fn}$;\\
\STATE Sort the detection results $B_d$ in descending order according to their corresponding detection scores $S_d$; 
\FOR{$i<N_d$}
\FOR{$j<N_g$}
\STATE Compute the overlap $O_j$ between the bounding-box $B_d^i$ and the ground-truth $B_g^j$;
\ENDFOR
\STATE Compute maximum overlap $O_m=\max\limits O_j$ and corresponding index $j_m=\arg\max O_j$; 
\IF{$O_m>\alpha$}
\STATE Add the corresponding $B_{d}^{i}$ to the set $B_{tp}$;
\STATE Remove $B_{d}^{i}$ and $B_{g}^{j_m}$ from $B_{d}$ and $B_{g}$;
\ELSE
\STATE Add the corresponding $B_{d}^{i}$ to the set $B_{fp}$;
\STATE Remove $B_{d}^{i}$ from $B_{d}$;
\ENDIF
\ENDFOR
\STATE Add $B_g$ to $B_{fn}$;
\RETURN $B_{tp}$, $B_{fp}$, $B_{fn}$;
\end{algorithmic}
\end{algorithm}
\subsection{Multispectral pedestrian datasets}
Multispectral pedestrian datasets in Table \ref{table:datasets} (bottom) are built on visible-light and thermal cameras. Thus, more useful information is provided for  robust pedestrian detection.

\textbf{KAIST}\footnote{\url{https://soonminhwang.github.io/rgbt-ped-detection}} \cite{Hwang_KAIST_CVPR_2015} is a large-scale multispectral pedestrian dataset recorded by a specitically designed imaging hardware device, which can capture the aligned color and thermal image pairs. This dataset has 95,328 image pairs with 103,128 dense annotations and 1,182 unique pedestrians.

\textbf{CVC-14}\footnote{\url{http://adas.cvc.uab.es/elektra/datasets}} \cite{Gonzalez_CVC_Sensors_2016} is a dataset of multimodal (FIR and visible) video sequences recorded during daytime and nighttime. There are 7,085 images for training and 1,433 images for testing. Different from KAIST dataset, the alignment is mainly achieved through post-processing, since the  resolution and the field-of-view of two sensors are different.

\subsection{Evaluation metrics}
Three evaluation metrics, \textit{i.e.,} log-average miss rate (MR), average precision (AP), and jaccard index (JI), are typically used in pedestrian detection. Among these, MR and AP are widely used, whereas JI was recently introduced for crowded pedestrian detection. Before discussing these evaluation metrics in more detail, we first explain how to determine if a detected bounding-box is a true positive or a false positive. The overlap between a detected bounding-box $B_d$ and a ground-truth $B_g$ can be calculated as
\begin{equation}
O=\frac{B_d \cup B_g}{B_d \cap B_g}.
\end{equation}
If the overlap $O$ is larger than a threshold of $\alpha$, the detected bounding-box is a potential matching with the ground-truth. Since a detected bounding-box might match multiple ground-truths, a greedy matching strategy in Algorithm \ref{Alg:GreedyMatch} splits the detection results into true positives and false positives along with the generated false negatives (missed positives). The threshold $\alpha$ is usually set to be 0.5. Based on true positives, false positives, and false negatives on the whole test set, log-average miss rate and average precision can be calculated to compare the detector performance.

\textbf{Log-average miss rate} Given a threshold of detection confidence score,  miss rate (MR) can be calculated by the number of true positives ($N_{tp}$) and the number of ground-truths ($N_g$) as 
\begin{equation}
MR = 1-N_{tp} / N_{g},
\end{equation}
and false positives per image (FPPI) can be calculated by dividing false positives by the number of images.
By varying detection confidence threshold, miss rates against false positives per image (FPPI) can be plotted in log-space. Finally, the log-average miss rate MR is calculated by averaging miss rates under 11 FPPI equally spaced in [$10^{-2}$:$10^{0}$].  A lower MR reflects a better performance.

The Caltech \cite{Dollar_Caltech_PAMI_2010} and CityPersons \cite{Zhang_Citypersons_CVPR_2017} datasets evaluate MR under different sets, \textit{e.g.,} \textbf{R}, \textbf{RS}, \textbf{HO}, \textbf{R+HO}, and \textbf{A}. The \textbf{R} set comprises pedestrians over 50 pixels in height with less than 35\% occlusion. The \textbf{RS} set  comprises pedestrians over 50 pixels and under 75 pixels with less than 0.35 occlusion.
The \textbf{HO} set comprises pedestrians over 50 pixels in height with 35-80\% occlusion. The \textbf{R+HO} set is the union set of \textbf{R} and \textbf{HO}. The \textbf{A} set contains the pedestrians over 20 pixels in height with less than 80\% occlusion.

\textbf{Average precision} Given a threshold of confidence score, recall ($R$) can be calculated by the number of true positives ($N_{tp}$) and the number of ground-truths ($N_g$) as
\begin{equation}
R = N_{tp} / N_{g}.
\end{equation}
The precision ($P$) can be calculated by the number of true positives ($N_{tp}$) and the number of all detected bounding-boxes ($N_d$) as
\begin{equation}
P = N_{tp} / N_{d}.
\end{equation}
By varying the threshold of confidence score, precision against recall can be plotted as a curve. Based on the precision-recall curve, the average precision (AP) is calculated by averaging precisions under 41 recalls equally spaced in [0:1]. A higher AP reflects better performance.

The KITTI \cite{Geiger_KITTI_CVPR_2012} dataset evaluates AP under three sets, \textit{i.e.,} \texttt{Easy}, \texttt{Medium}, and \texttt{Hard}. The \texttt{Easy} set includes pedestrians over 40 pixels in height with no occlusion.  The \texttt{Medium} set includes pedestrians over 25 pixels in height with less than part occlusion.  The \texttt{Hard} set includes pedestrians over 25 pixels in height with less than heavy occlusion.

\textbf{Jaccard index.} The recently proposed Jaccard index (JI) represents the overlap between detection results and ground-truths \cite{Shao_CrowdHuman_arXiv_2018}, which is used to evaluate the performance in crowded scenes. The Jaccard index score $S_{JI}$ under a given confidence threshold can be computed as: 
\begin{equation}
    S_{JI} = \frac{\text{IoUMatch}(\mathcal{D},\mathcal{G})}{| \mathcal{D} | + |\mathcal{G} | - |\text{IoUMatch}(\mathcal{D} ,\mathcal{G})|},
\end{equation}
where $\mathcal{D}$ represents the set of detection results, and $\mathcal{G}$ is the set of ground-truths. 
$\mathcal{M_D}, \mathcal{M_G} = \text{IoUMatch}(\mathcal{D} ,\mathcal{G})$, where $\mathcal{M_D}$ and $\mathcal{M_G}$ 
are maximum matching detection sets and ground truth sets, respectively. The IoU match is computed using the Hungarian algorithm. When increasing the confidence threshold, the JI score first increases and then deceases. To generate the best JI score for a detector, the greedy searching algorithm is used to compare the JI scores under different confidence thresholds. If the detector has a larger best JI score, the detector has a better performance.

\begin{table}[t]
\centering
\caption{Miss rates (MR) of 30 state-of-the-art methods on Caltech pedestrian dataset. The sets of \textbf{R}, \textbf{HO}, \textbf{R+HO}, and \textbf{A} are used for evaluation. The top part is based on the standard annotations of Caltech \cite{Dollar_Caltech_PAMI_2010}, and the bottom part is based on the new and accurate annotations \cite{Zhang_HowFar_CVPR_2016}.  The superscripts 
C/K/X/Z/1 indicate the CPU, NVIDIA K40 GPU, NVIDIA TitanX GPU, NVIDIA TitanZ GPU, and NVIDIA 1080Ti GPU.}
\resizebox{0.475\textwidth}{!}{
\setlength\tabcolsep{4pt}
\renewcommand\arraystretch{1.1}
\begin{tabular}{|l|c|c|c|c|c|c|c|}  
\hline
Method & Family & Backbone & Time   &\textbf{R}$\downarrow$  &\textbf{HO}$\downarrow$ &\textbf{R+HO}$\downarrow$ &\textbf{A}$\downarrow$\\
\hline
\hline
ACF \cite{Dollar_ACF_PAMI_2014} & CF & - & 0.11$^C$ & 44.2 & 90.2 & 54.6 & 79.6\\
LDCF \cite{Nam_LDCF_NIPS_2014} & CF & - & 0.28$^C$ & 24.8  & 81.3 &  37.7 & 71.2\\
Katamari \cite{Benenson_TYPD_ECCV_2014} & DPM &  -  &  - & 22.5  & 84.4 &  36.2 & 71.3\\
DeepCascade \cite{Angelova1_DeepCascade_BMVC_2015} & Hybrid & - & 0.67$^C$ & 31.1  & 81.7 & 42.4 & 74.1\\
SCCPriors \cite{Yang_SCCPriors_BMVC_2015} & CF & - & 3.88$^C$ & 21.9  & 80.9 &  35.1 & 70.3\\
TA-CNN \cite{Tian_TACNN_CVPR_2015} & Hybrid &  AlexNet & - &  20.9 & 70.4 &  33.3 & 71.2\\
CCF \cite{Yang_CCF_ICCV_2015} & Hybrid & AlexNet & 10.0$^Z$ &  18.7 & 72.4 &  30.6 & 66.7\\
Checkerboards \cite{Zhang_FCF_CVPR_2015} & CF & - & 2.0$^X$ & 18.5  & 77.5 &  31.8 & 68.7\\
DeepParts \cite{Tian_DeepParts_ICCV_2015} & Hybrid &  GoogleNet & - & 11.9 & 60.4 & 22.8 & 64.8\\
CompACT-Deep \cite{Cai_CompACT_ICCV_2015}& Hybrid & VGG16 & 0.5$^K$ & 11.7 & 65.8 & 24.6 & 64.4\\
MS-CNN \cite{Cai_MSCNN_ECCV_2016} & P-CNN & VGG16 & 0.13$^X$ & 10.0 & 59.9 & 21.5 & 60.9\\
RPN+BF \cite{Zhang_RPN+BF_ECCV_2016} & Hybrid & VGG16 &  0.5$^K$ & 9.6 & 74.3 & 24.0 & 64.7\\
F-DNN \cite{Du_FDNN_WACV_2017} & P-CNN & GoogleNet & 0.3$^X$ & 8.6 & 55.1 & 19.3 & 50.6\\
PCN \cite{Wang_PCN_BMVC_2017} & P-CNN & VGG16 & - & 8.4 & 55.8 & 19.2 & 61.9\\
PDOE \cite{Zhou_Bibox_ECCV_2018} & P-CNN & VGG16 & - & 7.6 & 44.4 & - & -\\
UDN+ \cite{Ouyang_UDN_PAMI_2018} & P-CNN & VGG16 & - & 11.5 & 70.3 & 24.7 & 64.8\\
FRCNN+ATT \cite{Zhang_FasterATT_CVPR_2018} & P-CNN & VGG16 & - & 10.3 & 45.2 & 18.2 & 54.5\\
SAF-RCNN \cite{Li_SARCNN_TMM_2018} & Hybrid & VGG16 & 0.59$^X$ & 9.7 & 64.4 & 21.9 & 62.6\\
ADM \cite{Zhang_ADM_TIP_2018} & P-CNN & ResNet50& - & 8.6 & 30.4 & 13.7 & 42.3\\
GDFL \cite{Lin_GDFL_ECCV_2018} & P-CNN &  VGG16 & 0.05$^1$ & 7.8 & 43.2 & 15.6 & 48.1\\
TLL-TFA \cite{Song_TLL_ECCV_2018} & P-CNN & ResNet50& - & 7.4 & \textbf{28.7} & \textbf{12.3} & \textbf{38.2}\\
AR-Ped \cite{Brazil_ANP_CVPR_2019} & P-CNN & VGG16 & 0.09$^1$ & \textbf{6.5} & 48.8 & 16.1 & 58.9\\
FRCN+A+DT \cite{Zhou_DFT_ICCV_2019} & P-CNN & VGG16 & - & 8.0 & 37.9 & - & -\\
MGAN \cite{Pang_MGAN_ICCV_2019} & P-CNN & VGG16 & - & 6.8 & 38.1 & 13.8 & -\\
\hline
\hline
HyperLearner \cite{Mao_HyperLearner_CVPR_2017} &  P-CNN &  VGG16 &  - & 5.5 & - &- &-\\
RepLoss \cite{Wang_Repulsion_CVPR_2018} &  P-CNN &  ResNet50&  - & 4.0 & - &- &-\\
ALFNet \cite{Liu_ALFNet_ECCV_2018} &  P-CNN &  ResNet50& 0.05$^1$ & 4.5 & - &- &-\\
OR-CNN \cite{Zhang_ORCNN_ECCV_2018} &  P-CNN &  VGG16  & - & 4.1 & - &- &-\\
JointDet \cite{Chi_JointDet_AAAI_2020} &  P-CNN &  ResNet50& - & 3.0 & - &- &-\\
PedHuter \cite{Chi_PedHunter_AAAI_2020} &  P-CNN &  ResNet50& - & \textbf{2.3} & - &- &-\\
\hline
\end{tabular}
}
\label{table:caltech}
\end{table}

\subsection{State-of-the-art comparison and analysis}
Here, we provide a comparison and discussion of several state-of-the-art methods on four widely used datasets (\textit{i.e.,} Caltech \cite{Dollar_Caltech_PAMI_2010}, KITTI \cite{Geiger_KITTI_CVPR_2012}, CityPersons \cite{Zhang_Citypersons_CVPR_2017}, and KAIST \cite{Hwang_KAIST_CVPR_2015}). Besides these state-of-the-art comparison, we provide more comprehensive analysis about pedestrian detection.

Table \ref{table:caltech} compares some methods on Caltech pedestrian dataset \cite{Dollar_Caltech_PAMI_2010}. 
Four subsets of \textbf{R}, \textbf{HO}, \textbf{R+HO}, and \textbf{A} are used for performance evaluation. 
Among these methods, only 5 approaches (\textit{i.e.,} ACF \cite{Dollar_ACF_PAMI_2014}, LDCF \cite{Nam_LDCF_NIPS_2014}, Katamari \cite{Benenson_TYPD_ECCV_2014}, SCCPriors \cite{Yang_SCCPriors_BMVC_2015}, and Checkerboards \cite{Zhang_FCF_CVPR_2015}) are handcrafted features based methods. The remaining methods are deep features based approaches. Miss rates witnessed a significant drop due to the introduction of deep methods DeepParts \cite{Tian_DeepParts_ICCV_2015} and CompaACT-Deep \cite{Cai_CompACT_ICCV_2015}. On \textbf{R} set, the best method is two-stage  AR-Ped \cite{Brazil_ANP_CVPR_2019}. On \textbf{HO}, \textbf{R+HO}, and \textbf{A} sets, the best method is TLL-TFA \cite{Song_TLL_ECCV_2018}, which uses time-sequence information for detection.  A likely reason is that time-sequence information plays an important role in occluded and small-scale pedestrian detection. 

By using the new and accurate annotations \cite{Zhang_HowFar_CVPR_2016} of Caltech pedestrian dataset, some state-of-the-art methods (\textit{e.g.,} JointDet \cite{Chi_JointDet_AAAI_2020} and PedHunter \cite{Chi_PedHunter_AAAI_2020}) report a relatively lower miss-rate on \textbf{R} set. These two methods use the head information to improve pedestrian detection. Further, the lower miss-rate indicates that the performance on Caltech pedestrian dataset is close to be saturated.

\begin{table}[t]
\centering
\caption{Average precisions (AP) of 21 state-of-the-art methods on the KITTI test set. The superscripts 
C/K/X/1 indicate the CPU, NVIDIA K40 GPU, NVIDIA TitanX GPU, and NVIDIA 1080Ti GPU.}
\resizebox{0.475\textwidth}{!}{
\setlength\tabcolsep{6pt}
\renewcommand\arraystretch{1.1}
\begin{tabular}{|l|c|c|c|c|c|c|}  
\hline
Method & Family & Backbone  & Time   &\texttt{Medium}$\uparrow$  &\texttt{Easy}$\uparrow$&\texttt{Hard}$\uparrow$\\
\hline
\hline
ACF \cite{Dollar_ACF_PAMI_2014} & CF  & - & 0.2$^C$ & 39.81 & 44.49 & 37.21\\
Checkerboards \cite{Zhang_FCF_CVPR_2015} & CF  & - & 2.0$^C$ & 56.75  & 67.65 & 51.12 \\
DeepParts \cite{Tian_DeepParts_ICCV_2015} & Hybrid  & GoogleNet & 1.0$^C$ & 58.67 & 70.49 & 52.78\\
CompACT-Deep \cite{Cai_CompACT_ICCV_2015}& Hybrid  & VGG16 & 1.0$^K$ & 58.74 & 70.69 & 52.71\\
Regionlets \cite{Wang_Regionlets_PAMI_2015} & DPM  & - & 1.0$^C$& 60.83 & 73.79 & 54.72\\
NNNF \cite{Cao_NNNF_CVPR_2016} & CF   & - & 0.5$^C$ & 58.01  & 69.16 & 52.77 \\
RPN+BF \cite{Zhang_RPN+BF_ECCV_2016}& Hybrid   & VGG16 & 0.6$^K$ & 61.29 & 75.45 & 56.08\\
SDP+RPN \cite{Yang_SDP_CVPR_2016} & Hybrid   & VGG16 & 0.4$^K$ & 70.42 & 82.07 &  65.09\\
IVA \cite{Zhu_SADRN_ACCV_2016} & P-CNN   & VGG16 & 0.4$^X$ & 71.37 & 84.61 & 64.90\\
MS-CNN \cite{Cai_MSCNN_ECCV_2016} & P-CNN   & VGG16 & 0.4$^X$ & 74.89 & 85.71 & 68.99\\
SubCNN \cite{Xiang_SACNN_WACV_2017} & P-CNN  & GoogleNet & - & 72.27 & 84.88 & 66.82\\
GN \cite{Jung_DNGN_PRL_2017} & Hybrid   & VGG16 & 1.0$^X$ & 72.29 & 82.93 & 65.56\\
RRC \cite{Ren_RRC_CVPR_2017} & P-CNN  & VGG16 & 3.6$^X$ & \textbf{76.61} & 85.98 & \textbf{71.47}\\
CFM \cite{Hu_CFM_CSVT_2018}&  CF   & VGG16 & 2.0$^K$ & 62.84& 74.76 & 56.06\\
SJTU-HW \cite{Zhang_LED_ICIP_2018} & P-CNN   & VGG16 & 0.85$^X$ & 75.81  & 87.17 & 69.86\\
GDFL \cite{Lin_GDFL_ECCV_2018} & P-CNN   & VGG16 & 0.27$^1$ & 68.62 & 84.61 & 66.86\\
MonoPSR \cite{Ku_MonoPSR_CVPR_2019}& P-CNN    & ResNet101 & 0.2$^X$  & 68.56 & 85.60 & 63.34\\
FFNet \cite{Zhao_MPOEL_PR_2019} & P-CNN    & VGG16 & 1.07$^1$ & 75.99  & \textbf{87.21} & 69.86\\
MHN \cite{Cao_MHN_TCSVT_2019} & P-CNN   & VGG16  & 0.39$^X$  & 75.99  & \textbf{87.21} & 69.50\\
Aston-EAS \cite{Wei_EOD_TITS_2019} & P-CNN  & VGG16 & 0.24$^1$ & 76.07 & 86.71 & 70.02\\
AR-Ped \cite{Brazil_ANP_CVPR_2019} & P-CNN   & VGG16 & - & 73.44 & 83.66 & 68.12\\
\hline
\end{tabular}}
\label{table:kitti}
\end{table}

\begin{table}[t]
\centering
\caption{Miss rates (MR) of 17 state-of-the-art methods on CityPersons validation set. The superscript 
1 indicates the NVIDIA 1080Ti GPU. The superscript $^\dagger$ indicates the pedestrians over 50 pixels in height with more than 35\% occlusion, instead of pedestrians over 50 pixels in height with 35-80\% occlusion. Thus, $^\dagger$ suggest higher difficulty.}
\resizebox{0.475\textwidth}{!}{
\setlength\tabcolsep{8pt}
\renewcommand\arraystretch{1.1}
\begin{tabular}{|l|c|c|c|c|c|c|}  
\hline
Method & Family & Backbone     & Scale & Time &\textbf{R}$\downarrow$ &\textbf{HO}$\downarrow$\\
\hline
\hline
Adapted FR-CNN \cite{Zhang_Citypersons_CVPR_2017} & P-CNN & VGG16 & 1$\times$ & - & 15.4 & - \\
RepLoss \cite{Wang_Repulsion_CVPR_2018} & P-CNN & ResNet50 & 1$\times$ & -  & 13.7 & 56.9$^\dagger$ \\
FRCNN+ATT \cite{Zhang_FasterATT_CVPR_2018} & P-CNN & VGG16 & 1$\times$ & - & 16.0 & 56.7 \\
TLL+MRF \cite{Song_TLL_ECCV_2018}  & P-CNN & ResNet50& 1$\times$ &- & 14.4 & 52.0$^\dagger$ \\
OR-CNN \cite{Zhang_ORCNN_ECCV_2018} & P-CNN & VGG16 & 1$\times$ & -  & 12.8 & 55.7$^\dagger$ \\
ALFNet \cite{Liu_ALFNet_ECCV_2018} & P-CNN & ResNet50 & 1$\times$ & 0.27$^1$ & 12.0 & 51.9$^\dagger$ \\
CSP \cite{Liu_CSP_CVPR_2019} & P-CNN & ResNet50 & 1$\times$ & 0.33$^1$ & 11.0 & 49.3$^\dagger$ \\
Adaptive-NMS \cite{Liu_AdaptiveNMS_CVPR_2019} & P-CNN& VGG16 & 1$\times$& -& 11.9 & 55.2$^\dagger$\\
MGAN \cite{Pang_MGAN_ICCV_2019} & P-CNN & VGG16 & 1$\times$ & -& 11.3 & \textbf{42.0} \\
R$^2$NMS \cite{Huang_PBM_CVPR_2020}& P-CNN & VGG16  & 1$\times$ & -& 11.1 & \textbf{53.3}$^\dagger$ \\
PRNet \cite{Song_PRN_ECCV_2020}& P-CNN & ResNet50 & 1$\times$ & 0.22$^1$ & \textbf{10.8} & \textbf{42.0} \\
\hline
\hline
Adapted FR-CNN \cite{Zhang_Citypersons_CVPR_2017}  & P-CNN & VGG16 & 1.3$\times$ &- & 12.8 & - \\
RepLoss \cite{Wang_Repulsion_CVPR_2018} & P-CNN & ResNet50 & 1.3$\times$ & -& 11.6 & 55.3$^\dagger$ \\
OR-CNN \cite{Zhang_ORCNN_ECCV_2018}& P-CNN & VGG16 & 1.3$\times$ &- & 11.0 & 51.3$^\dagger$ \\
PDOE \cite{Zhou_Bibox_ECCV_2018} & P-CNN & VGG16  & 1.3$\times$ & - & 11.2 & 44.2 \\
Adaptive-NMS \cite{Liu_AdaptiveNMS_CVPR_2019}  & P-CNN  & VGG16  & 1.3$\times$  & - & 10.8 & 54.2$^\dagger$ \\
IoU$_{vis}$+Sign \cite{Lu_VBSP_ICIP_2019} & P-CNN & VGG16  &1.3$\times$ & - & 10.8 & 54.3$^\dagger$ \\
FRCN+A+DT \cite{Zhou_DFT_ICCV_2019} & P-CNN & VGG16  & 1.3$\times$ & - & 11.1 & 44.3 \\
MGAN \cite{Pang_MGAN_ICCV_2019} & P-CNN & VGG16 & 1.3$\times$ & - & 10.5 & \textbf{39.4} \\
JointDet \cite{Chi_JointDet_AAAI_2020} & P-CNN & ResNet50& 1.3$\times$ & -& 10.2 & - \\
0.5-stage \cite{Ujjwal_OHS_WACV_2020} & P-CNN & ResNet50& 1.3$\times$ & - & \textbf{8.1} & - \\
PedHunter \cite{Chi_PedHunter_AAAI_2020} & P-CNN & ResNet50& 1.3$\times$& -& 8.3  & \textbf{43.5}$^\dagger$ \\
\hline
\end{tabular}}
\label{table:citypersons}
\end{table}
Table \ref{table:kitti} compares several state-of-the-art methods on the KITTI test set \cite{Geiger_KITTI_CVPR_2012}. The methods only using 2D image annotations are selected for fair comparison. Among these these methods, only four methods (\textit{i.e.,} ACF \cite{Dollar_ACF_PAMI_2014}, Checkerboards \cite{Zhang_FCF_CVPR_2015}, NNNF \cite{Cao_NNNF_CVPR_2016}, and Regionlets \cite{Wang_Regionlets_CVPR_2013,Wang_Regionlets_PAMI_2015}) are handcrafted features based approaches and the remaining methods are deep features based approaches.   On \texttt{Medium} and \texttt{Hard} sets, RRC \cite{Ren_RRC_CVPR_2017} and Aston-EAS \cite{Wei_EOD_TITS_2019} are the top two methods. On \texttt{Easy} set, FFNet \cite{Zhao_MPOEL_PR_2019} and MHN \cite{Cao_MHN_TCSVT_2019} are the top two methods. Most of these methods adopt  feature pyramid structure with multi-scale feature fusion and data augmentation strategy (\textit{e.g.,} multi-scale training). Compared with the \texttt{Easy} set, the \texttt{Hard} set contains more small-sized pedestrians, occluded pedestrians, and truncated pedestrians. As a result, the \texttt{Hard} set provides more than 10\% lower performance, which indicates that the small-sized  and occluded pedestrian detection are the two main bottlenecks.

\begin{table}[t]
\centering
\caption{Miss rates of state-of-the art detectors on CityPersons test sets.}
\resizebox{0.475\textwidth}{!}{
\setlength\tabcolsep{5pt}
\renewcommand\arraystretch{1.1}
\begin{tabular}{|l|c|c|cccc|}
\hline
Method	 & Family & Backbone & \textbf{R}$\downarrow$	&\textbf{RS}$\downarrow$	&\textbf{HO}$\downarrow$ &\textbf{A}$\downarrow$ \\\hline\hline
MS-CNN  \cite{Cai_MSCNN_ECCV_2016}    & P-CNN &    VGG16   & 13.32 & 15.86 & 51.88 & 39.94 \\ 
Adapted FR-CNN \cite{Zhang_Citypersons_CVPR_2017} & P-CNN & VGG16 & 12.97 & 37.24 & 50.47 & 43.86 \\
Cascade MS-CNN \cite{Cai_Cascade_CVPR_2018} &  P-CNN & VGG16 & 11.62 & 13.64 & 47.14 & 37.63 \\
Repulsion Loss \cite{Wang_Repulsion_CVPR_2018} & P-CNN &  ResNet50& 11.48 & 15.67 & 52.59 & 39.17 \\
Adaptive-NMS \cite{Liu_AdaptiveNMS_CVPR_2019}  &  P-CNN &  VGG16  & 11.40 & 13.64 & 46.99 & 38.89 \\
OR-CNN  \cite{Zhang_ORCNN_ECCV_2018}     &    P-CNN & VGG16  & 11.32 & 14.19 & 51.43 & 40.19 \\
MGAN   \cite{Pang_MGAN_ICCV_2019}       &    P-CNN & VGG16 & 9.29  & 11.38 & 40.97 & 38.86 \\
APD \cite{Zhang_APD_arXiv_2019}          &   P-CNN & DLA34 & 8.27  & 11.03 & 35.45 & 35.65 \\
Pedestron  \cite{Hasan_EIR_arXiv_2020}   & P-CNN & HRNet  & 7.69  & 9.16  & 27.08 & 28.33\\
\hline
\end{tabular}}
\label{table:citypersons-test}
\end{table}

\begin{table}[t]
\centering
\caption{Miss rates of state-of-the art detectors on KAIST \textbf{R} test set using the annotations provided by \cite{Liu_MPNN_BMVC_2016}. The all test set (All) contains the day subset (Day) and the night subset (Night). The superscripts X/1 indicate NVIDIA Titan X/1080Ti.}
\resizebox{0.475\textwidth}{!}{
\setlength\tabcolsep{5pt}
\renewcommand\arraystretch{1.1}
\begin{tabular}{|l|c|c|c|ccc|}
\hline
 Method	 &  Family &  Backbone&  Time	&   All$\downarrow$	&  Day$\downarrow$ &  Night$\downarrow$ \\\hline\hline
 Halfway Fusion \cite{Liu_MPNN_BMVC_2016}  & Hybrid &   VGG16     &  0.43$^X$ &  25.75 &   24.88 &   26.59 \\
 Fusion RPN \cite{Konig_FusionRPN_CVPR_2017} & Hybrid &   VGG16      &  0.80$^X$ &  25.75 &   24.88 &   26.59\\
 IAFR-CNN \cite{Li_IAFRCNN_PR_2019}      & P-CNN &  VGG16    &  0.21$^X$ &  15.73 &   14.55 &   18.26 \\
 IATDNN+IASS   \cite{Guan_IADNN_IF_2019}    & P-CNN   &    VGG16   &  0.25$^X$  &  14.95	&  14.67	&  15.72 \\
 CIAN  \cite{Zhang_CMIAN_IF_2019}    & P-CNN  &   VGG16   &  0.07$^1$  &  14.12 &  14.77 &  11.13 \\
 MSDSR-CNN \cite{Li_MSDS_BMVC_2018}  & P-CNN  &   VGG16   &  0.23$^X$  &  11.34 &   10.53 &  12.94\\
 AR-CNN \cite{Zhang_ARRCNN_ICCV_2019}    & P-CNN     &   VGG16   &  0.12$^1$  &  9.34 &  9.94 & 	 8.38 \\
 MBNet  \cite{Zhou_MBNet_ECCV_2020} & P-CNN  & ResNet50    & 0.07$^1$  & \textbf{8.13} &	\textbf{8.28} &	\textbf{7.86} \\
\hline
\end{tabular}}
\label{table:kaist}
\end{table}

Table \ref{table:citypersons} compares several state-of-the-art methods on CityPersons validation set \cite{Zhang_Citypersons_CVPR_2017}. All these methods are deep features based methods. Two subsets of \textbf{R} and \textbf{HO}, which adopt similar metrics as that of the Caltech dataset \cite{Dollar_Caltech_PAMI_2010}, are used here for performance evaluation. Note that there are two different settings about \textbf{HO}. Most of these state-of-the-art methods are two-stage methods, and are the variants of Faster R-CNN \cite{Ren_FasterRCNN_NIPS_2015}. Additionally, 0.5-stage detector \cite{Ujjwal_OHS_WACV_2020} that uses the pseudo-segmentation for anchor generation also achieves the state-of-the-art performance. For occluded pedestrian detection, the methods (\textit{i.e.,} MGAN \cite{Pang_MGAN_ICCV_2019}, JointDet\cite{Chi_JointDet_AAAI_2020}, PedeHutter \cite{Chi_PedHunter_AAAI_2020}, and R$^2$NMS \cite{Huang_PBM_CVPR_2020}) using part information (\textit{e.g.,} visible and head annotations) have a better performance. Table \ref{table:citypersons-test} further shows miss-rates of several state-of-the-art methods on CityPersons test set \cite{Zhang_Citypersons_CVPR_2017}.

In addition to detection accuracy, we also provide the test time of different methods in the three datasets above. By comprehensive analysis over the three datasets, the single-stage methods usually have faster speed. For example, GDFL \cite{Lin_GDFL_ECCV_2018} and ALFNet \cite{Liu_ALFNet_ECCV_2018} are superior in terms of speed on Caltech test set, while PRNet \cite{Song_PRN_ECCV_2020} and CSP \cite{Liu_CSP_CVPR_2019} are the fastest on CityPersons validation set. In addition, these single-stage methods have comparable accuracy to state-of-the-art two-stage methods. This suggests that  single-stage methods usually have a better trade-off between accuracy and detection speed.

Table \ref{table:kaist} further compares some state-of-the-art methods on multispectral KAIST test set \cite{Hwang_KAIST_CVPR_2015}, including hybrid methods and pure CNN based methods. With illumination aware feature alignment and single-stage structure, MBNet \cite{Zhou_MBNet_ECCV_2020} achieves the best performance on both speed and accuracy.

\begin{table}[t]
\centering
\caption{Improving generic object detector for pedestrian detection on the CityPersons validation set.}
\resizebox{0.46\textwidth}{!}{
\setlength\tabcolsep{5pt}
\renewcommand\arraystretch{1.1}
\begin{tabular}{|l|c|ccc|}
\hline
 Method	&  Backbone	&   \textbf{R}$\downarrow$	&  \textbf{HO}$\downarrow$ & \textbf{A}$\downarrow$ \\\hline\hline
FPN-vanilla \cite{Lin_FPN_CVPR_2017}   &  ResNet50 & 18.11 &   50.99 &  43.11 \\
+ignored region handling     &  ResNet50&  14.81 &  48.09 &   39.88\\
+anchor aspect ratios \& scales        &  ResNet50& 14.26 &  44.52 &   38.17 \\
\hline
\end{tabular}}
\label{table:obj2ped}
\end{table}

\begin{table}[t]
\centering
\caption{Impact of extra-class training on pedestrian detection. Simply using additional object classes cannot improve pedestrian detection.}
\resizebox{0.46\textwidth}{!}{
\setlength\tabcolsep{5pt}
\renewcommand\arraystretch{1.1}
\begin{tabular}{|l|c|ccc|}
\hline
Dataset	&  Training strategy 	&  \textbf{R}$\downarrow$	&  \textbf{HO}$\downarrow$ &  \textbf{A}$\downarrow$ \\\hline
\multirow{2}*{TJU-Ped-Traffic \cite{Pang_DHD_TIP_2020}}   &  Multi-class training &  22.59 &   61.23 &   38.36\\
    &  Single-class training &  22.36 &   60.47 &   37.78\\
\hline
\hline
Dataset	&  Training strategy	&   AP$\uparrow$	&  AP@0.5$\uparrow$ &  AP@0.75$\uparrow$ \\\hline
\multirow{2}*{MS COCO  \cite{Lin_COCO_ECCV_2014}}  &  Multi-class training &  52.2 &   81.5 &   56.5\\ 
 &  Single-class training &  53.2 &  81.9 &  57.2\\ 
\hline
\end{tabular}}
\label{table:multiclass}
\end{table}

\begin{table}[t]
\centering
\caption{Impact of different data processing operations during training on the CityPersons validation set.}
\resizebox{0.45\textwidth}{!}{
\setlength\tabcolsep{5pt}
\renewcommand\arraystretch{1.1}
\begin{tabular}{|l|c|ccc|}
\hline
Method	&  Backbone	&  \textbf{R}$\downarrow$	&  \textbf{HO}$\downarrow$ & \textbf{A}$\downarrow$ \\\hline\hline
(a) Baseline (FPN)   &  ResNet50& 14.26 &   44.52 &   38.17\\
(b) Reasonable pedestrians only   &  ResNet50&  14.34 &  57.71 &   45.10\\ 
(c) Multi-scale training    &  ResNet50&  13.36 &  44.02 &   35.99 \\
(d) Random erase augmentation        &  ResNet50&  13.97 &  43.35 &   37.31 \\
(e) Copy-paste augmentation        &  ResNet50&  14.10 &   43.92 &  38.30 \\
\hline
\hline
(f) Caltech$\rightarrow$CityPersons       &  ResNet50& 13.92 &   45.56 &   38.35 \\
(g) MS COCO$\rightarrow$CityPersons   &  ResNet50&  11.96 &  39.43 &  35.10\\
(h) CrowdHuman$\rightarrow$CityPersons   &  ResNet50&  11.62 &   38.13 &  34.23\\ 
(i) TJU-Pedestrian$\rightarrow$CityPersons     &  ResNet50&  9.80 &   34.68 &   34.51 \\
\hline
\end{tabular}}
\label{table:dataped}
\end{table}
\begin{table}[t]
\centering
\caption{Impact of different backbones on the CityPersons validation set.}
\resizebox{0.45\textwidth}{!}{
\setlength\tabcolsep{5pt}
\renewcommand\arraystretch{1.1}
\begin{tabular}{|l|c|c|ccc|}
\hline
Method	&  Backbone&  AP on COCO $\uparrow$ 	&   \textbf{R}$\downarrow$	& \textbf{HO}$\downarrow$ &  \textbf{A}$\downarrow$ \\\hline\hline
(a) FPN  &  ResNet50&  37.4 &  14.26 &   44.52 &  38.17\\
(b) FPN   &  VGG16 &  - &  13.79 &  44.10 &   39.51\\ 
(c) FPN   &  ResNet101 &  39.4 &  14.55 &   44.10 &   39.03\\ 
(d) FPN        &  RegNet &  39.9 &  13.29 &   45.82 &  37.31 \\
(e) FPN     &  HRNet &  40.2 &  12.64 &   41.02 &  35.19\\
\hline
\end{tabular}}
\label{table:datafeat}
\end{table}

\begin{table}[t]
\centering
\caption{Impact of different NMS strategies on the CityPersons validation set.}
\resizebox{0.45\textwidth}{!}{
\setlength\tabcolsep{5pt}
\renewcommand\arraystretch{1.1}
\begin{tabular}{|l|c|c|ccc|}
\hline
Method	&  Backbone	& Threshold 	& \textbf{R}$\downarrow$	& \textbf{HO}$\downarrow$ & \textbf{A}$\downarrow$ \\\hline\hline
\multirow{3}*{(a) NMS}    &  ResNet50&  $\theta=0.4$ & 14.67 &   45.46 &   38.69\\
   &  ResNet50&  $\theta=0.5$ &  14.26 &   44.52 &   38.17\\
   &  ResNet50&  $\theta=0.6$ &  14.46 &   44.01 &   39.05\\   
\hline
\multirow{3}*{(b) SoftNMS}      &  ResNet50&  $\theta=0.4$&  14.02 &  44.76 &   38.44 \\ 
      &  ResNet50&  $\theta=0.5$&  14.11 &  44.25 &   38.12 \\ 
      &  ResNet50&  $\theta=0.6$&  14.46 &  43.98 &   39.05 \\
\hline
\end{tabular}}
\label{table:post}
\end{table}

\begin{table}[t]
\centering
\caption{Generalization ability of the detector on different datasets.}
\resizebox{0.45\textwidth}{!}{
\setlength\tabcolsep{5pt}
\renewcommand\arraystretch{1.1}
\begin{tabular}{|l|c|cc|}
\hline
Method	&  Train 	&   CityPersons $\downarrow$	& Caltech $\downarrow$  \\\hline\hline
FPN \cite{Lin_FPN_CVPR_2017}   &  Caltech & 57.1 &  11.0 \\ 
Cascade R-CNN \cite{Cai_Cascade_CVPR_2018}   & Caltech &  57.8 &   10.7 \\ 
\hline
FPN \cite{Lin_FPN_CVPR_2017}   &  CityPersons &  14.3 &   24.9 \\
Cascade R-CNN \cite{Cai_Cascade_CVPR_2018}   & CityPersons &  13.8 &   24.2 \\ 
\hline
FPN \cite{Lin_FPN_CVPR_2017}   & CityPersons+Caltech &  16.5 &   12.6 \\ 
Cascade R-CNN \cite{Cai_Cascade_CVPR_2018}   &  CityPersons+Caltech &  16.5 &  11.8\\ 
\hline
\end{tabular}}
\label{table:crossdomain}
\end{table}

To give more insights on pedestrian detection, we further analyze pedestrian detection from different aspects, including comparison with generic object detector, the impact of data processing, the impact of feature extraction, the impact of post-processing, and generalization ability analysis.

\textbf{Comparison with generic object detector} Here, we provide an analysis regarding using a generic object detector FPN for pedestrian detection in Table \ref{table:obj2ped}. We perform the 1$\times$ training scheme (12 epochs) on two GPUs with the initial learning rate of 0.005. There are two images per GPU. All the pedestrians over 20 pixels in height with less than 80\% occlusion are used for training. The original FPN achieves 18.11\% MR on \textbf{R} set. By ignoring the proposals located in ignored regions, it provides 3.30\% improvement on \textbf{R} set. When further setting the aspect ratio as 0.41 and uniform scale, it provides 0.55\% improvement. With these simple modifications specific  to pedestrian detection, our improved FPN has a comparable, yet not state-of-the-art, performance in pedestrian detection. Finally, our improved FPN is also used as the baseline detector in the following experiments.

Generally, the generic object detector simultaneously detects multiple object classes. A natural question to ask is: does pedestrian detection benefit from additional object classes. Table \ref{table:multiclass} shows the impact of extra-class training on pedestrian detection. Experiments are performed on two datasets, including TJU-Ped-Traffic \cite{Pang_DHD_TIP_2020} and MS COCO \cite{Lin_COCO_ECCV_2014}. TJU-Ped-Traffic \cite{Pang_DHD_TIP_2020} has 5 object classes, while MS COCO \cite{Lin_COCO_ECCV_2014} has 80 object classes. 
We observe that performing multi-class training does not improve over single-class training. We believe how to effectively exploit the contextual information and the relation between person and other objects is an interesting future direction.

\textbf{Impact of data processing}  Here, we analyze the impact of different data processing on pedestrian detection in Table \ref{table:dataped}. The first (top) half of Table \ref{table:dataped} shows the impact of different intra-dataset data processing strategies. (1) Instead of the pedestrians over 20 pixels in height with less than 80\% occlusion for training, we use the pedestrians over 50 pixels in height with less than 65\% occlusion (see (b)). We observe MR on the \textbf{R} set to be similar, but MRs on \textbf{HO} and \textbf{A} sets have a large drop. This shows that adding occluded and small-scale pedestrians during training is important for improving occluded and small-scale pedestrian detection. (2) Using multi-scale training strategy  (see (c)) is effective for improving pedestrian detection. (3) We also randomly erase the part of pedestrians with pixel mean value during the training (see (d)), observing improvement in occluded pedestrian detection likely due to enhancing the diversity of pedestrian occlusion. (4) To enlarge the number of pedestrians, we randomly paste the pedestrians (bounding-box) from other images with geometric prior. We observe no significant improvement in performance likely due to the fact that a simple bounding-box copy-paste also incorporates some background information and brings the style inconsistency (\textit{e.g.,} illumination variation).

The bottom part of Table \ref{table:dataped} shows the impact of pre-training on different different datasets, including Caltech \cite{Dollar_Caltech_PAMI_2010}, MS COCO \cite{Lin_COCO_ECCV_2014}, CrowdHuman \cite{Shao_CrowdHuman_arXiv_2018}, and TJU-Pedestrian \cite{Pang_DHD_TIP_2020}. When performing pre-training on Caltech, we observe no significant improvement in performance on the \textbf{R} set.  The large-scale generic object dataset MS COCO provides 2.3\% improvement on \textbf{R} set. However, pre-training with MS COCO is inferior to the person related datasets, namely CrowdHuman \cite{Shao_CrowdHuman_arXiv_2018} and TJU-Pedestrian \cite{Pang_DHD_TIP_2020}. Among these person related datasets, pre-training on TJU-Pedestrian leads to more favorable performance on \textbf{R} set, likely due to a large number of pedestrians and the related scenes (pedestrian scenarios).

\textbf{Impact of feature extraction}  Second, we show the impact of feature extraction using different backbones in Table \ref{table:datafeat}, including ResNet50 \cite{He_ResNet_CVPR_2016}, ResNet101 \cite{He_ResNet_CVPR_2016}, VGG16 \cite{Simonyan_VGG_arXiv_2014}, RegNet \cite{radosavovic2020designing}, and HRNet \cite{Wang_HRNet_PAMI_2020}. Among these backbones, HRNet achieves the best detection performance, especially on \textbf{HO} and \textbf{A}. The reason maybe explained as that  high-resolution and high-semantic representations are important for improving pedestrian detection.

\textbf{Impact of post-processing}  Third, we show the impact of two post-processing strategies (\textit{i.e.,} NMS and SoftNMS) in Table \ref{table:post}. It is difficult to achieve the optimal results on all the sets by using a single threshold $\theta$. Some recent methods \cite{Chi_PedHunter_AAAI_2020,Huang_PBM_CVPR_2020} adopt visible part or head for occluded pedestrian detection, which not only need extra annotations but also face the problem of threshold settings. Thus, it is interesting to pay more attention on NMS-free methods in future.

\textbf{Generalization ability}  Finally, we analyze generalization ability by performing cross-dataset evaluations in Table \ref{table:crossdomain}. Both FPN and Cascade R-CNN have a sub-optimal performance during cross-dataset evaluation (see rows 1-4), which indicates poor generalization ability. Similar findings are also presented in CSP \cite{Liu_CSP_CVPR_2019}. Furthermore, we analyze the effect of the combined training on CityPersons and Caltech datasets (see rows 5-6). In both cases, we observe that the combined training does not achieve better performance, compared to individual dataset-specific training.

\section{Challenges}
\label{sec-challenges}
Despite this great success, pedestrian detection still faces several challenges discussed as follows. 

\begin{figure}[t]
\centering
\includegraphics[width=0.99\linewidth]{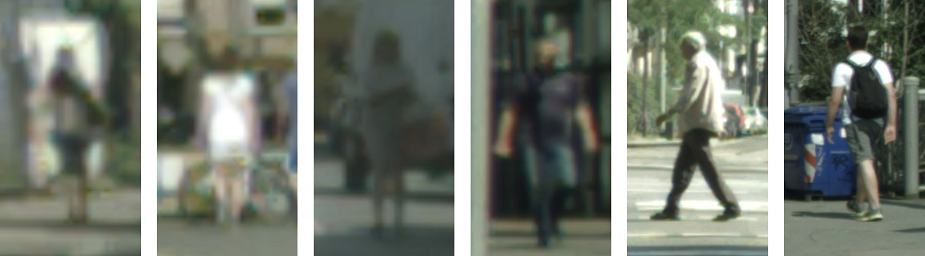}
\caption{Example pedestrians of various scales. From left to right, pedestrians vary from small scale to large scale. Pedestrians of different scales have large-scale variations and small-scale pedestrians are relatively noisy and blurry.}
\label{fig:challengeSmall}
\end{figure}

\subsection{Scale variance}
Traffic and video surveillance scenes usually contain  pedestrians of various scales. Fig. \ref{fig:challengeSmall} shows some examples. Large-scale and small-scale pedestrians exhibit high intra-class variations. As a result, it is challenging to use a single detector to accurately detect pedestrians of varying scales.

One of straightforward idea is scale independent strategy, including image pyramid based and feature pyramid based methods. For image pyramid based methods, it is important to improve its efficiency. In contrast, feature pyramid strategy is relatively effective. However, how to extract robust scale-independent features is important for feature pyramid strategy. Another idea is reducing the difference between pedestrians of different scales \cite{Pang_JCSNet_TIFS_2019,Wu_SML_ACM_2020,Yin_MRGAN_ICIP_2019}. We can exploit to reduce intra-class variations in both data and feature levels with generative adversarial networks.

In addition, small-scale pedestrian detection is the bottleneck for solving scale variance problem. However, the existing methods lack of the enough research on small scale object detection \cite{Yu_SMA_WACV_2020,Han_SSPD_TITS_2019}. Thus, it it necessary to treat small-scale pedestrian detection as a standalone problem.

\begin{figure}[t]
\centering
\includegraphics[width=0.99\linewidth]{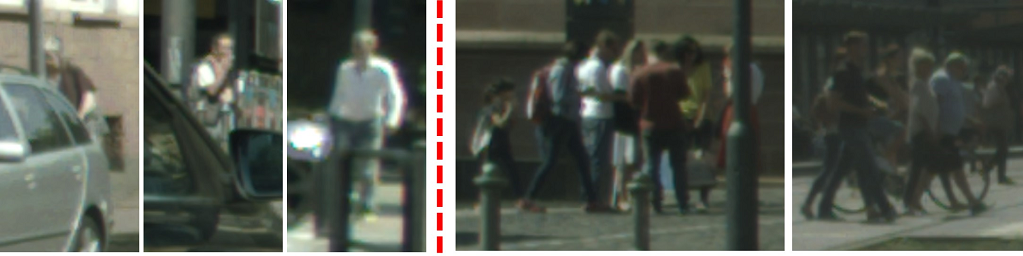}
\caption{Example pedestrians under different types of occlusion (\textit{i.e.,} inter-class and intra-class occlusions). Some examples of inter-class occlusion are shown in the left, where the level of occlusion varies from heavy to bare. Some examples of intra-class occlusion are shown in the right, where intra-class occlusion occurs between different pedestrians.}
\label{fig:challengeOcc}
\end{figure}

\subsection{Occlusion}
Pedestrian occlusion is a very common problem. For instance, 70\% of pedestrians in CityPersons dataset \cite{Zhang_Citypersons_CVPR_2017} are occluded. Fig. \ref{fig:challengeOcc} shows two types of pedestrian occlusions: inter-class occlusion and intra-class occlusion. Inter-class occlusion occurs when pedestrians are occluded by other objects (not pedestrians). In contrast, intra-class occlusion occurs when pedestrians are occluded by other pedestrians.

The main solution to inter-class occlusion is enhancing the features of unoccluded part and suppressing the features of occluded part. There are two strategies: implicit strategy \cite{Tian_DeepParts_ICCV_2015,Xie_PSC_arXiv_2020} and explicit strategy \cite{Zhou_Bibox_ECCV_2018,Pang_MGAN_ICCV_2019,Chi_PedHunter_AAAI_2020}. The implicit strategy usually learns multiple part detectors to cover different occlusion patterns. In this strategy, how to set/combine the parts is an open  problem. The explicit strategy uses extra annotations (\textit{e.g.,} head or visible-part annotation). Because the extra annotations need large resource consumption, it is useful to exploit using a small number of part annotations to help occluded pedestrian detection.

Usually, pedestrian detection needs to remove duplicate bounding-boxes by NMS. As a negative effect, NMS combines the bounding-boxes belonging to different pedestrians in crowd scene (intra-class occlusion).  To solve this problem, one solution is the improved NMS strategy, for example, dynamic NMS threshold. Another solution is NMS-free strategy by avoiding bounding-box combination operation.

Besides,  temporal information is important for occluded pedestrian detection in driving and surveillance scenes.

\subsection{Domain adaptation}
Most existing methods focus on a specific pedestrian dataset and can not guarantee a good domain adaptation ability  \cite{Wang_SSPD_PAMI_2014,Wang_FARCNN_CVPR_2019}. For instance, the pedestrian detector trained under good weather often has a sub-optimal performance in poor weather (\textit{e.g.,} fog and rain). Therefore, it is necessary to address the issue of domain adaptation. Most domain adaptation methods are based on adversarial learning, including data-level \cite{Kim_DM_CVPR_2019,Hsu_PDA_CVPRW_2019,Khodabandeh_RLA_ICCV_2019,Lopez_DASC_BMVC_2019}, feature-level \cite{Saito_SWDA_CVPR_2019,Xie_MDAL_ICCVW_2019}, and instance-level \cite{Zhu_AOD_CVPR_2019,Chen_DARCNN_CVPR_2018} methods. However, they mainly focus on generic object detection. In future, the domain adaptation methods about pedestrian detection can be exploited by considering pedestrian/scene characteristic.

Besides, we  can exploit both the same-domain and cross-domain evaluations in future, which guides the researchers to design the pedestrian detector with good comprehensive performance.

\subsection{Multi-sensor fusion}
In Section \ref{sec-mspd}, multispectral pedestrian detection adopts two kinds of sensors (\textit{i.e.,} visible-light/infrared cameras). It provides more robust detection performance for illumination variance. To ensure safety and generate 3D information, we not only need to fuse information from multiple sensors of homogeneous data (visible-light/infrared cameras) but also multiple sensors of heterogeneous data (cameras and LiDAR). LiDAR can provide accurate depth information while cameras has detailed semantic information. However, how to fuse information in heterogeneous data is a challenging and important task in future.

\subsection{Real-time detection}
Most existing pedestrian detection methods focus on improving detection accuracy, while ignoring the efficiency. However, the application to driving/surveillance scenes has limited computational resources but requires real-time detection speed.  For example, the fastest method reported on CityPersons has 0.22s inference speed on a single NVIDIA 1080Ti GPU, which can't meet the needs of real applications. Thus, it is necessary to study the light-weight and real-time pedestrian detection methods for the embedded device.

\section{Conclusion}

In past decade, pedestrian detection has witnessed significant success, which has gone from the handcrafted features to deep features based approaches. In this paper, we first summarize these two types of methods in detail. Afterwards, we review multispectral pedestrian detection. We review popular pedestrian datasets and a deep analysis on pedestrian detection methods. Finally, we discuss some challenging problems (\textit{e.g.,} occlusion, scale variance, and domain adaptation) in pedestrian detection. We hope that this deep survey can help the researchers to develop new methods in the field of pedestrian detection.


%





\ifCLASSOPTIONcaptionsoff
  \newpage
\fi



%


\bibliographystyle{ieee_fullname}
\bibliography{refs}

%


\begin{IEEEbiography}[{\includegraphics[width=1in,height=1.25in,clip,keepaspectratio]{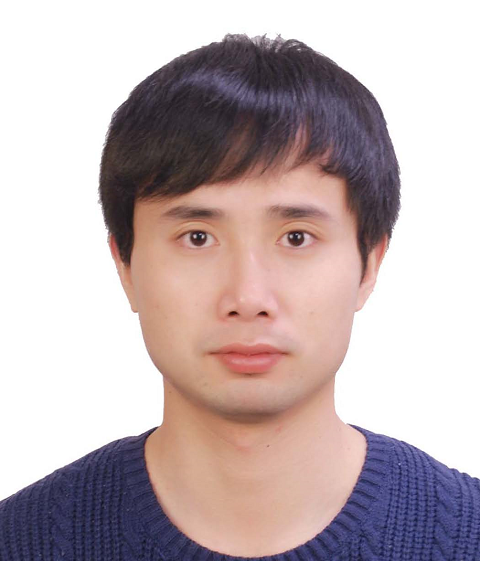}}]{Jiale Cao} received the Ph.D. degree in information and communication engineering from the Tianjin University, Tianjin, China, in 2018. He is currently an associate professor at the Tianjin University. His research interests include object detection and deep learning, in which he has published 15 paper in CVPR, ICCV, ECCV, IEEE TIP, IEEE TCSVT and IEEE TIFS.
\end{IEEEbiography}

\begin{IEEEbiography}[{\includegraphics[width=1in,height=1.25in,clip,keepaspectratio]{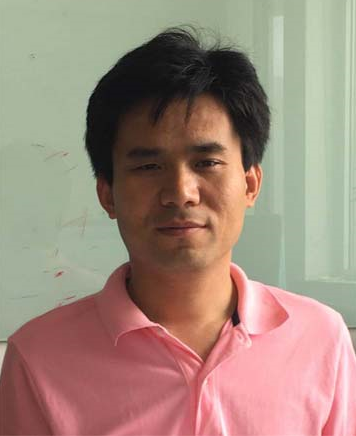}}]{Yanwei Pang} received the Ph.D. degree in electronic engineering from the University of Science and Technology of China. Currently, he is a professor at the Tianjin University and is also the founding director of the Tianjin Key Laboratory of Brain Inspired Intelligence Technology (BIIT). He has published 150 scientific papers including 70 papers in top journals/conferences. He is an associate editor of both IEEE T-NNLS and Neural Networks (Elsevier).
\end{IEEEbiography}

\begin{IEEEbiography}[{\includegraphics[width=1in,height=1.25in,clip,keepaspectratio]{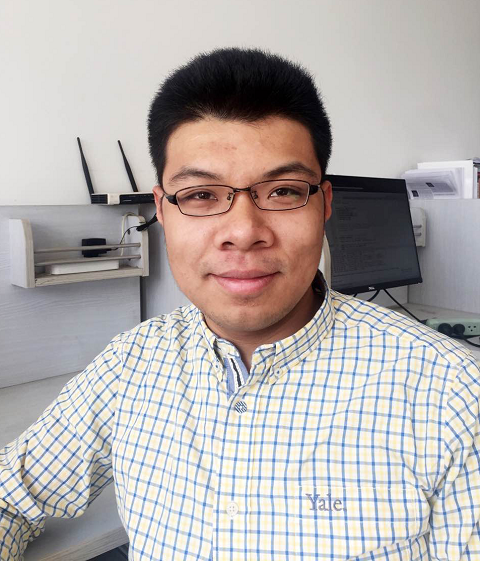}}]{Jin Xie} received the B.S. degree in electronic engineering from the Tianjin University, Tianjin, China, in 2016. He is currently pursuing the Ph.D. degree in the Tianjin University and his supervisor is Prof. Yanwei Pang. His research interests include machine learning and computer vision, in which he has published 5 paper in CVPR, ICCV, ECCV, IEEE TCSVT, IEEE TCYB.
\end{IEEEbiography}

\begin{IEEEbiography}[{\includegraphics[width=1in,height=1.25in,clip,keepaspectratio]{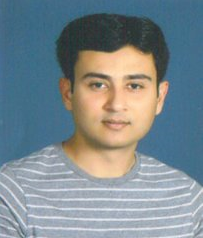}}]{Fahad Shahbaz Khan}  received the Ph.D. degree in Computer Vision from Computer Vision Center Barcelona and Autonomous University of Barcelona, Spain. He is a faculty member at Mohamed bin Zayed University of Artificial Intelligence, UAE and Link{\"o}ping University, Sweden.  His research interests include a wide range of topics within computer vision. 
\end{IEEEbiography}

\begin{IEEEbiography}[{\includegraphics[width=1in,height=1.25in,clip,keepaspectratio]{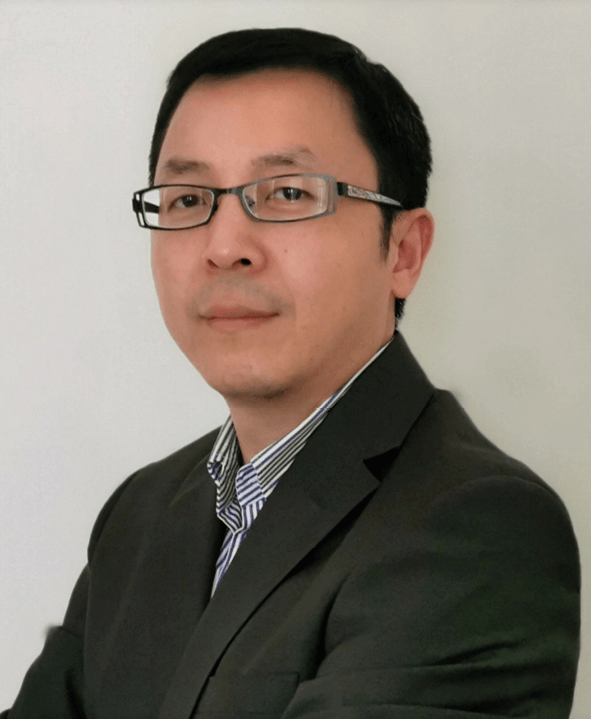}}]{Ling Shao} is the CEO and the Chief Scientist of the Inception Institute of Artificial Intelligence (IIAI), Abu Dhabi, United Arab Emirates. His research interests include computer vision, machine learning, and medical imaging. He is a fellow of the IEEE, the IAPR, the IET, and the BCS.  
\end{IEEEbiography}






\end{document}